\documentclass[10pt,twocolumn,letterpaper]{article}

\usepackage{cvpr}              
\usepackage{graphicx}
\usepackage{amsmath}
\usepackage{amssymb}
\usepackage{booktabs}
\usepackage{colortbl}
\usepackage{multirow}
\usepackage{soul}
\usepackage{adjustbox}
\usepackage{pifont} 
\usepackage{tabularx}
\usepackage{makecell}
\newcommand{\cmark}{\textcolor[rgb]{0.502, 0.788, 0.643}{\textbf{\checkmark}}}
\newcommand{\xmark}{\textcolor[rgb]{0.800, 0.447, 0.541}{\textbf{\ding{55}}}}
\usepackage{amsthm}

\newtheorem{proposition}{Proposition}

\definecolor{cvprblue}{rgb}{0.21,0.49,0.74}
\usepackage[pagebackref,breaklinks,colorlinks,allcolors=cvprblue]{hyperref}

\def\paperID{***} 
\def\confName{CVPR}
\def\confYear{2025}

\title{Multi-Level Collaboration in Model Merging}
\author{
    Qi Li, Runpeng Yu, Xinchao Wang\thanks{Corresponding Author.}\\
    National University of Singapore \\
    {\tt\small \{liqi, r.yu\}@u.nus.edu, xinchao@nus.edu.sg}
}

\begin{document}
\maketitle
\begin{abstract}
Parameter-level model merging is an emerging paradigm in multi-task learning with significant promise. Previous research has explored its connections with prediction-level model ensembling—commonly viewed as the upper bound for merging—to reveal the potential of achieving performance consistency between the two. However, this observation relies on certain preconditions, such as being limited to two models, using ViT-based models, and all models are fine-tuned from the same pre-trained checkpoint. To further understand the intrinsic connections between model merging and model ensembling, this paper explores an interesting possibility: If these restrictions are removed, can performance consistency still be achieved between merging and ensembling? To answer this question, we first theoretically establish a performance correlation between merging and ensembling. We find that even when previous restrictions are not met, there is still a way for model merging to attain a near-identical and superior performance similar to that of ensembling. To verify whether our findings are practical, we introduce a validation framework termed Neural Ligand (NeuLig). The learning process of NeuLig is meticulously designed with a specialized loss function supported by theoretical foundations. Experimental results demonstrate the robust resilience of NeuLig in terms of both model scale and the number of collaborating models. For instance, for the case involving 5 CLIP-ViT-B/32 models, parameter-level merging achieves the same performance as prediction-level ensembling (merging: 95.44\% vs. ensembling: 95.46\%). Please check our repo \href{https://github.com/LiQiiiii/Neural-Ligand}{here}.
\end{abstract}
\vspace{-3mm}

\section{Introduction}
\label{sec:intro}
Exploring the intrinsic connections between multiple models to enable efficient reuse and collaboration has long been a core issue in multi-task learning \cite{zhang2021survey,chen2024multi, li2024encapsulating, li2023towards}. Recently, a new paradigm named model merging \cite{wortsman2022model,ilharcoediting,yadav2024ties,jin2023dataless,yangadamerging,tangmerging} has emerged. This paradigm merges single-task models at the parameter-level using task vectors \cite{ilharcoediting} to create a unified multi-task model. 
It enables the seamless integration of specialized knowledge from each single-task model, resulting in enhanced efficiency and adaptability across tasks.

Besides model merging, another widely-used paradigm in multi-task learning called model ensembling \cite{dong2020survey,sagi2018ensemble,polikar2012ensemble,zhou2021ensemble,zhou2021domain,zhang2012ensemble} improves multi-task learning performance by aggregating outputs from multiple models rather than merging parameters. Recent studies \cite{wortsman2022robust,jacot2018neural,fort2020deep,wortsman2022model} have started to explore the connections between these two paradigms. For example, results in \cite{wortsman2022model} suggest that when two models collaborate, parameter-level merging and prediction-level ensembling can yield nearly comparable performance at the data level. Despite the promising results, prior related work primarily focuses on collaborations that (1) center around ViT-based models, (2) are fine-tuned from the same pre-trained checkpoint, and (3) are limited to only two models. 

We argue that a deeper investigation into this phenomenon is both necessary and valuable, as current model merging methods—typically relying on task vectors or their variations—often yield performance that falls short of standalone models \cite{ilharcoediting,yadav2024ties,jin2023dataless,yangadamerging} and ensembling is often considered an upper bound on merging performance \cite{stoica2023zipit}. An open question remains as to whether it is theoretically feasible for merging to match the performance of ensembling in cases involving multiple models (more than two, which is the primary focus of merging) and, if so, under what specific conditions such consistency could be achieved. Achieving this can greatly liberate the potential of multi-model collaboration, especially when multiple entities collaborate while privacy constraints vary across stages—sometimes allowing model parameter sharing, while at other times only permitting model output sharing \cite{yao1982protocols, gentry2009fully, smith2017federated}. Grounded in this significance, we take an initial step toward exploring this intriguing question: \textit{When multiple models collaborate, can we find an inherent link between parameter-level merging and prediction-level ensembling that could enable performance consistency for both sides?} We first theoretically demonstrate that merging and ensembling can yield a performance difference of the second order of smallness under certain conditions. This holds regardless of model scale and the number of collaborating models.

\begin{figure}[t]
  \centering
  \includegraphics[width=0.9\linewidth]{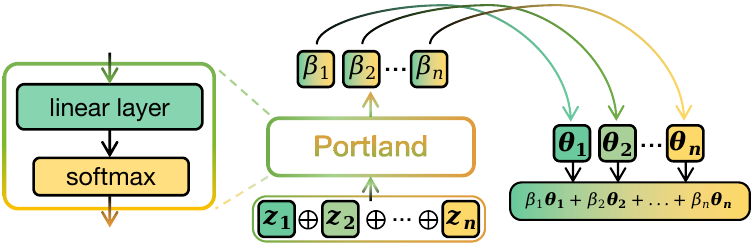}
   \caption{An illustration of Portland, which consists of a linear layer followed by a softmax function.}
   \vspace{-5mm}
   \label{fig:portland}
\end{figure}
Furthermore, to verify it in practice, we design a validation framework called \texttt{\textbf{\underline{Neu}ral \underline{Lig}and} (NeuLig)}. In \texttt{NeuLig}, we train an extremely small (single-layer) neural network termed Portable Ligand (Portland). As shown in Figure \ref{fig:portland}, Portland takes the outputs of several models on each data as input and generates a corresponding Cooperative Vector (CoopVec), with one entry per model for each data. These vectors can then be used for either parameter-level merging or prediction-level ensembling. The learning process of Portland is meticulously designed with a specialized loss function, which is backed by our theoretical findings. The loss function comprises two terms: the boosting term and the alignment term. We adopt an ensemble-driven merging strategy, where the boosting term guides Portland to enhance prediction-level ensembling performance, while 
the alignment term is designed to bring the performance of merging in line with that of ensembling. Ultimately, this enables the realization of performance consistency between merging and ensembling. Results on multi-model collaboration (e.g., five CLIP-RN-50 models or five CLIP-ViT-B/32 models) demonstrate the effectiveness of \texttt{NeuLig}. Furthermore, we find that the performance consistency persists even when models are trained from random initialization rather than fine-tuned from the same pre-trained checkpoint. We also explore the behavior of \texttt{NeuLig} across different scenarios and provide relevant insights.
Our contributions are summerized as follows:

\begin{itemize}[leftmargin=2em]
    \item We explore whether parameter-level merging and prediction-level ensembling can achieve performance consistency in multi-model collaboration at the data level, offering theoretical support for this potential.
    
    \item We develop a validation framework named \texttt{\textbf{\underline{Neu}ral \underline{Lig}and} (NeuLig)} to further verify our findings. In this framework, we introduce a single layer network termed Portland to perform merging and ensembling in a unified manner.
    
    \item Extensive experimental results validate the feasibility of \texttt{NeuLig}. For instance, we observe an almost negligible performance gap (0.02\%) between merging (95.44\%) and ensembling (95.46\%) in the collaboration of five ViT-based models. We further explore the nature of NeuLig from various perspectives and provide detailed discussions, offering a new perspective on understanding model merging and ensembling.
\end{itemize}
\section{Related Work}
\label{sec:related}
\noindent\textbf{Model Merging for Multi-Task Learning.}
Model merging has received significant attention for its storage- and computation-efficient properties, showing promise in improving model generalization and supporting multi-task learning (MTL) \cite{zhang2021survey,evgeniou2004regularized,sener2018multi, li2023towards}. From a task scope perspective, current merging methods can be divided into two categories \cite{yangadamerging,zhang2024badmerging}: single-task and multi-task model merging. The former merges multiple models trained on the same task, either to improve generalization \cite{stochastic-weight-averaging-in-parallel,cha2021swad,wortsman2022model} or to enable federated learning \cite{Li2020On,Wang2020Federated,liu2022deep}. In contrast, multi-task model merging combines models from different tasks to perform MTL \cite{wortsman2022model,ilharcoediting,yadav2024ties,jin2023dataless,yangadamerging,tangmerging}. This line of work focuses on a broadly applicable multi-task scenario, for which many promising techniques have been developed. For example, Task Arithmetic \cite{ilharcoediting} introduces the concept of `task vectors', showing that merging these vectors to create a unified model can effectively support MTL. Based on the concept of task vector, Ties-Merging \cite{yadav2024ties} addresses task conflicts in Task Arithmetic by resetting redundant parameters, resolving sign conflicts, and selectively merging parameters that exhibit sign consistency. RegMean \cite{jin2023dataless} proposes minimizing the \(L_2\) distance between the merged model and each individual model. AdaMerging \cite{yangadamerging} highlights the crucial role of coefficients in the model merging process for achieving optimal performance, specifically addressing this factor to bridge the performance gap. Other router-based methods \cite{tangmerging,muqeethsoft} draw inspirations from Mixture-of-Expert paradigm \cite{shazeer2017outrageouslylargeneuralnetworks,zadouri2024pushing,hwang2024pre,lepikhin2020gshard}, trying to provide a dynamic operation mechanism for the merged model. In this paper, we focus on the multi-task model merging scenario.

\noindent\textbf{Correlation Between Merging and Ensembling.} Some works \cite{wortsman2022robust,jacot2018neural,fort2020deep,wortsman2022model} have started to study the connection between parameter-level merging and the well-established practice of prediction-level ensembling. \cite{wortsman2022robust} demonstrates that, under certain conditions, linearly combining the weights of a fine-tuned model with its original zero-shot model can approximate the effect of ensembling their predictions. This approximation holds when the loss can be locally expressed by a linear expansion, also known as the NTK regime \cite{jacot2018neural}. Furthermore, \cite{fort2020deep} finds that this linear approximation becomes increasingly accurate in the later stages of neural network training. When this approximation holds exactly, weight averaging and ensembling are equivalent \cite{wortsman2022robust}. \cite{wortsman2022model} gives a further analysis on this connection empirically, verifying the performance relationship between parameter-level merging and prediction-level ensembling in ViT-based models. In this paper, we extend the exploration of the performance relationship between merging and ensembling to more realistic scenarios, analyzing the properties and behaviors of both sides when applied to multiple models.

\label{sec:formatting}

\section{Method}
\label{sec:method}
In this section, we first give our theoretical findings for reducing the performance gap between model merging and ensembling. Following that, we introduce \texttt{\textbf{\underline{Neu}ral \underline{Lig}and} (NeuLig)}, a validation framework designed to assess the practical feasibility of our findings.

\subsection{Notations}
\begin{figure}[t]
  \centering
  \includegraphics[width=0.9\linewidth]{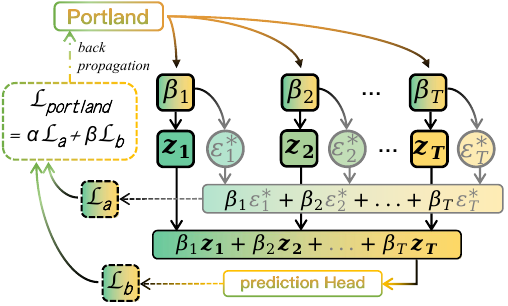}
   \caption{The training process of Portland. The CoopVec is combined separately with the model output and the modified offsets, contributing to two respective terms in the loss function.}
   \label{fig:portmand_train}
   \vspace{-5mm}
\end{figure}
Let $f_{\boldsymbol{\theta}}(\cdot)$ denote the output of a neural network (e.g., a visual encoder) parametrized by $\boldsymbol{\theta}$. Assuming that $\forall \boldsymbol{\theta}\in \Theta$, $f_{\boldsymbol{\theta}}(\cdot)$ is continuous and $\forall (x, y) \in \mathcal{D}$, $f_{\boldsymbol{\theta}}(x, y)$ is at least twice differentiable. For a group of \(T\) models parameterized by \(\boldsymbol{\theta_t} \in \Theta\) (\(t = 1,2,\dots,T\)), we define the average parameter vector \(\boldsymbol{\tilde{\theta}} = \sum_{t=1}^{T} \beta_t \boldsymbol{\theta_t}, \text{s.t. } \sum_{t=1}^{T} \beta_t = 1\), where \(\boldsymbol{\beta}=[\beta_1, \beta_2, \dots, \beta_T]\) represents the Cooperative Vector (CoopVec). Since we apply the same CoopVecs on both merging and ensembling, we use the unified notation \(\boldsymbol{\beta}\). The offset \(\boldsymbol{\xi_t}\) for each model's parameters is defined as \(\boldsymbol{\xi_t} = \boldsymbol{\theta_t} - \boldsymbol{\tilde{\theta}}\). For each query data \((x,y)\), we define the outcome for prediction-level ensembling as \(\tilde{f}(x,y) = \sum_{t=1}^{T}\beta_t f_{\boldsymbol{\theta_{t}}}(x,y)\), and the prediction after parameter-level merging is denoted as \(f_{\boldsymbol{\tilde{\theta}}}(x,y)\).

\subsection{Theoretical Discussion} 
\label{theory}
Now our primary objective is to verify whether merging and ensembling can achieve performance consistency in multi-model collaboration scenarios. Specifically, for any specific data \((x,y)\), we aim to validate whether:
\begin{equation}
\tilde{f}(x,y) - f_{\boldsymbol{\tilde{\theta}}}(x,y) \approx 0
\label{eq:diff_f}
\end{equation}
has an approximate solution that can be achieved under specific conditions to enable performance consistency. Since the model parameters of \( f_{\boldsymbol{\tilde{\theta}}}(x, y) \) are derived from all \( f_{\boldsymbol{\theta_t}}(x, y) \), it is natural for us to use a Taylor expansion to fit a quadratic polynomial of $f_{\boldsymbol{\tilde{\theta}}}(x, y)$ to approximate the value of each $f_{\boldsymbol{\theta_t}}(x, y)$:
\setlength{\abovedisplayskip}{5pt} 
\setlength{\belowdisplayskip}{5pt} 
\begin{align}
f_{\boldsymbol{\theta_t}}(x, y) &= f_{\boldsymbol{\tilde{\theta}}}(x, y) + \boldsymbol{\xi_t^\top} \nabla_{\boldsymbol{\xi_t}} f_{\boldsymbol{\tilde{\theta}}}(x, y) \nonumber \\
&\quad + \frac{1}{2} \boldsymbol{\xi_t^\top} \nabla_{\boldsymbol{\xi_t}}^2 f_{\boldsymbol{\tilde{\theta}}}(x, y) \boldsymbol{\xi_t} + O(\Delta^n),
\label{eq:taler}
\end{align}
where the parameter offset $\boldsymbol{\xi_t}$ for each model represents a neighborhood in which the Taylor expansion approximates the function around any given point (i.e., $\tilde{\boldsymbol{\theta}}$) in terms of its value and derivatives. $O(\Delta^n)$ denotes the higher-order remainder term. For simplicity, we omit \((x, y)\) in the following derivations. Thus, the difference between the output of prediction-level ensembling and that of parameter-level merging is expressed as:
\setlength{\abovedisplayskip}{2pt}
\setlength{\belowdisplayskip}{2pt}
\begin{align}
\tilde{f} - f_{\boldsymbol{\tilde{\theta}}} &= 
\sum_{t=1}^{T} \beta_t f_{\boldsymbol{\theta_t}} - f_{\boldsymbol{\tilde{\theta}}} \nonumber \\
&= \sum_{t=1}^{T} \beta_t f_{\boldsymbol{\tilde{\theta}}} + \sum_{t=1}^{T} \beta_t \boldsymbol{\xi_t^\top} \nabla_{\boldsymbol{\xi_t}} f_{\boldsymbol{\tilde{\theta}}} + \sum_{t=1}^{T} \beta_t O(\Delta^2) - f_{\boldsymbol{\tilde{\theta}}} \nonumber \\
&= \sum_{t=1}^{T} (\beta_t \boldsymbol{\xi_t^\top}) \nabla_{\boldsymbol{\xi_t}} f_{\boldsymbol{\tilde{\theta}}} + O(\Delta^2),
\label{eq:simplified_diff_f}
\end{align}
where the right-hand side (RHS) of the first equality follows from the previous definition of \(\tilde{f}(x, y)\); the RHS of the second equality is based on the relationship between \(\tilde{f}(x, y)\) and \(f_{\boldsymbol{\tilde{\theta}}}(x, y)\) as derived from the Taylor expansion in Equation \ref{eq:taler}; and the RHS of the third equality follows from \(\sum_{t=1}^{T} \beta_t = 1\), which allows the first and last terms to cancel out. Thus, the difference between \(\tilde{f}(x, y)\) and \(f_{\boldsymbol{\tilde{\theta}}}(x, y)\) is of second-order smallness if and only if \(\sum_{t=1}^{T} (\beta_t \boldsymbol{\xi_t^\top}) = \boldsymbol{0}\). We summarize this conclusion in the following proposition:
\noindent\begin{proposition}
For \( T \) neural networks parameterized by \( \boldsymbol{\theta_t} \) (where \( t = 1, 2, \dots, T \) and \( \forall \boldsymbol{\theta}_t \in \Theta \)), assuming $f_{\boldsymbol{\theta_t}}(\cdot)$ is continuous and $\forall (x, y) \in \mathcal{D}$, $f_{\boldsymbol{\theta}}(x, y)$ is (at least) twice differentiable. The performance difference between the prediction-level ensembling \(\tilde{f}(x,y)\) and the parameter-level merging $f_{\tilde{\boldsymbol{\theta}}}(x, y)$ is of the second order of smallness if and only if $\sum_{t=1}^{T} (\beta_t \boldsymbol{\xi_t}^{\top}) = \boldsymbol{0}$.
\end{proposition}
Denoting the parameter offset of each model as \(\boldsymbol{\xi_t}=[\xi_t^1, \xi_t^2, ..., \xi_t^n]\), we can derive an alternative form based on \textbf{Proposition 1} when the condition for second-order smallness is satisfied:
\begin{equation}
[\sum_{t=1}^{T}\beta_t\xi_t^1, \sum_{t=1}^{T}\beta_t\xi_t^2, ..., \sum_{t=1}^{T}\beta_t\xi_t^n] = \boldsymbol{0}.
\label{eq:eq0}
\end{equation}
Given this result, each entry on the left-hand side (LHS) must individually equal zero to satisfy the condition. This implies that, for each dimension \(i \in \{1, 2, \dots, n\}\), \(\sum_{t=1}^{T} \beta_t \xi_t^i\) must sum to zero. We denote \(\hat{\xi_t} = \sum_{i=1}^n \xi_t^i\). Summing those entries on the LHS above yields the following expression:
\begin{align}
&\sum_{t=1}^{T}\beta_t\xi_t^1 + \sum_{t=1}^{T}\beta_t\xi_t^2 + ... + \sum_{t=1}^{T}\beta_t\xi_t^n \nonumber\\ 
&= \beta_1\sum_{i=1}^n\xi_1^i + \beta_2\sum_{i=1}^n\xi_2^i + ... + \beta_T\sum_{i=1}^n\xi_T^i \nonumber\\
&= \sum_{t=1}^T \beta_t \sum_{i=1}^n \xi_t^i = \sum_{t=1}^T \beta_t \hat{\xi_t},
\label{alter_define}
\end{align}
which indicates that satisfying the conditions of \textbf{Proposition 1} is equivalent to ensuring that the aggregated parameter offsets of all models involved in the collaboration have no first-order impact on the result. In other words, minimizing the weighted sum of offsets ensures that the output of prediction-level ensembling becomes indistinguishable from that of parameter-level merging under a first-order approximation. This alignment facilitates performance consistency in multi-model collaboration.

\subsection{Neural Ligand}
Based on our findings in Section \ref{theory}, we introduce our validation framework \texttt{NeuLig}. Within this framework, we use an extremely small (single-layer) neural network, termed Portable Ligand (Portland), to generate CoopVecs for each model on each data. As shown in Figure \ref{fig:portland}, Portland takes the outputs of several models as input and generates corresponding CoopVecs, with one entry per model for each data. As illustrated in Figure \ref{fig:portmand_train}, for ease of display, we use \(\boldsymbol{z_t}\) to denote the output of the \(t^{\textrm{th}}\) model.

The learning process of Portland is designed with a specialized loss function with two components. As discussed in Section \ref{sec:intro}, the first component, termed the boosting term (\(\mathcal{L}_b\)), guides Portland to enhance the performance of ensembling. The second component, called the alignment term (\(\mathcal{L}_a\)), aligns the performance of merging with that of ensembling. The alignment term (\(\mathcal{L}_a\)) is supported by our findings in Section \ref{theory}. The interaction between these two terms influences the training of Portland, ultimately resulting in the performance consistency between merging and ensembling. We now provide the specific definitions of these two terms.

\begin{figure}[t]
  \centering
  \includegraphics[width=\linewidth]{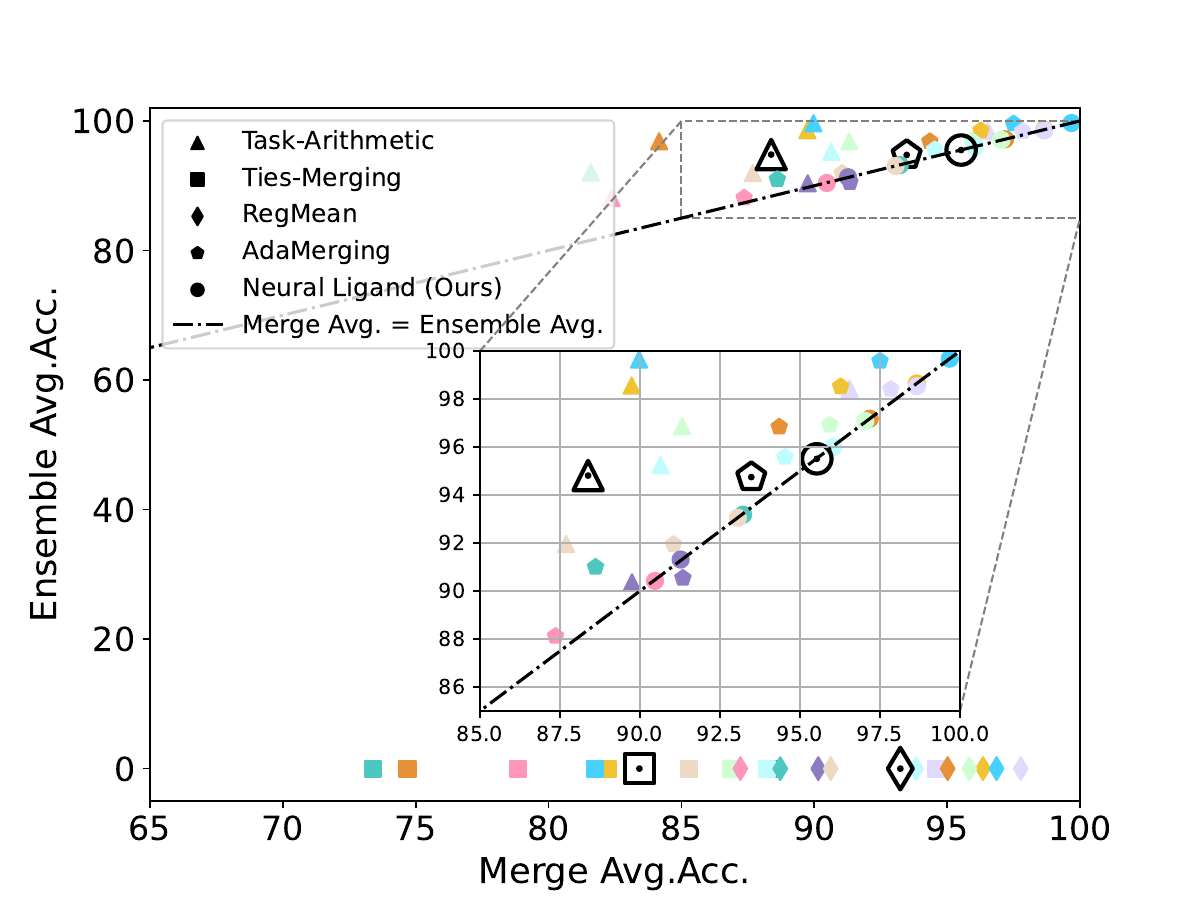}
   \caption{A toy experiment to verify theoretical feasibility. In this experiment, we merged two models that were fine-tuned on different datasets. \textbf{Marker shapes} represent different methods, while \textbf{colors} indicate different experimental groups, with each group using a distinct combination of datasets. In total, 10 groups are conducted (represented by 10 different colors). \textbf{Hollow markers} for each method indicate the average results across these 10 groups.}
   \label{fig:toy}
   \vspace{-3mm}
\end{figure}

\noindent\textbf{The Boosting Term (\(\mathcal{L}_b\)). }The purpose of this term is to ensure that the CoopVecs generated by Portland perform well in prediction-level ensembling. Following standard MTL setups, we consider two scenarios: supervised learning and semi-supervised learning. In the supervised setup, we assume that the original training sets used during the fine-tuning process of each model are available, enabling us to train Portland in a supervised manner using ground-truth labels. Specifically, as shown in Figure \ref{fig:portmand_train}, for any training data in the \(t^{\textrm{th}}\) model's training set, the \(\mathcal{L}_b\) term in the supervised scenario is defined as:
\begin{equation}
\mathcal{L}_{b}^{\textrm{sup}} = -\sum_{i=1}^{C_t} y_i \log(\sum_{t=1}^{T} \Psi_\phi(\boldsymbol{z_1}\oplus \boldsymbol{z_2}\oplus...\oplus \boldsymbol{z_T})_t \cdot \boldsymbol{z_t}),
\label{eq:sup_la}
\end{equation}
where \(\Psi_\phi(\cdot)_t\) is the \(t^{\textrm{th}}\) entry of the CoopVec generated by Portland, \(\oplus\) denotes a concatenation process, \(C_t\) is the total class numbers in the training set of the \(t^{\textrm{th}}\) model. In common model merging practice, the training set for each model is assumed not available. Thus, we further explore a semi-supervised setup. We draw inspirations from previous work \cite{yangadamerging} to use \textit{entropy minimization} \cite{grandvalet2004semi,roy2022uncertainty} on unlabeled test data as a surrogate objective function. \cite{yangadamerging} performs analysis on the correlation between entropy and prediction loss via the Spearman correlation coefficient and discovers a high positive correlation, thus view \textit{entropy minimization} as the optimization proxy goal in the semi-supervised setup. Here, for ease of display, we use \(\boldsymbol{\hat{z}}\) to denote the concatenated vector of each \(\boldsymbol{z_t}\) (\(\boldsymbol{\hat{z}}=\boldsymbol{z_1}\oplus \boldsymbol{z_2}\oplus...\oplus\boldsymbol{z_T})\). The \(\mathcal{L}_b\) term under semi-supervised setup can be written as:
\begin{small}
\begin{align}
\mathcal{L}_{b}^{\textrm{semi}} = -\sum_{i=1}^{C_t} [(\sum_{t=1}^{T} \Psi_\phi(\boldsymbol{\hat{z}})_t \cdot \boldsymbol{z_t}) \cdot \log(\sum_{t=1}^{T} \Psi_\phi(\boldsymbol{\hat{z}})_t \cdot \boldsymbol{z_t})].
\label{eq:semisup_la}
\end{align}
\end{small}
\noindent\textbf{The Alignment Term (\(\mathcal{L}_a\)).} The aim of this term is to align the utility of CoopVecs in parameter-level merging with that in prediction-level ensembling. Recall Equation \ref{alter_define}, we now use \(\boldsymbol{\xi_f}\) to denote \([\hat{\xi}_1, \hat{\xi}_2, ..., \hat{\xi}_T]\). To eliminate the impact of feature scaling and improve convergence speed during training\cite{hastie2009elements,han2022data}, we further apply scaling on \(\boldsymbol{\xi_f}\) using its standard deviation. Furthermore, to enhance training efficiency, we simplify the training process by pre-calculate an unweighted average of \(\boldsymbol{\tilde{\theta}}\) as an initialization on each training step. As shown in Figure \ref{fig:portmand_train}, we denote the final scaled \(\boldsymbol{\xi_f}\) as \(\boldsymbol{\xi_f^*} = [\xi^*_1, \xi^*_2, ..., \xi^*_T]\). The alignment term is defined as follows:
\begin{equation}
\mathcal{L}_{a} = \sum_{t=1}^{T} (\Psi_\phi(\boldsymbol{z_1}\oplus \boldsymbol{z_2}\oplus...\oplus\boldsymbol{z_T})_t \cdot \xi^*_t)^2,
\label{eq:lb}
\end{equation}
which can be viewed as a MSE loss on each data with a target value of 0. With the boosting and alignment terms defined above, as illustrated in Figure \ref{fig:portmand_train}, we define the final optimization objective for Portland as follows:
\begin{equation}
\mathcal{L}_{\textrm{Port}} = \alpha \mathcal{L}_{b} + \beta \mathcal{L}_{a},
\label{eq:final_conclusion}
\end{equation}
where \(\mathcal{L}_{b}\) can be either \(\mathcal{L}_{b}^{\textrm{sup}}\) or \(\mathcal{L}_{b}^{\textrm{semi}}\), \(\alpha\) and \(\beta\) are weighting coefficients that control the balance between the boosting term and the alignment term.

Before advancing to the experimental section, we conduct a preliminary verification with a toy experiment on \texttt{NeuLig}. As shown in Figure \ref{fig:toy}, this experiment involves collaboration between two CLIP-ViT-B/32 models (model related details are in Section \ref{mo_da}). We run a total of 10 experiments represented by different colors, each using different dataset combination (e.g., GTSRB-CIFAR10, RESISC45-CIFAR100, etc.). Additionally, we use various marker shapes to distinguish the results of different methods. Figure \ref{fig:toy} contains the results of prediction-level ensembling (y-axis) and parameter-level merging (x-axis) for different methods across these 10 experiments, as well as the average results for each method (indicated by hollow markers of the corresponding shape). In the zoomed-in view, the diagonal dashed line represents the results where exact performance consistency between ensembling and merging is achieved. Some of the baseline methods, such as Ties-Merging and RegMean, are not applicable for ensembling, so their value in the y-axis are set to zero. The results indicate that, for baseline methods applicable to both merging and ensembling, there is a noticeable performance gap, with ensembling generally outperforming merging. In contrast, \texttt{NeuLig} shows strong performance consistency between the two. Compared to baseline methods, the performance also shows a remarkable improvement. This provides initial evidence supporting \texttt{NeuLig}’s effectiveness as a validation framework. In the following section, we conduct more comprehensive experiments to further confirm its utility and share insights from these experiments.

\section{Experiments}
\label{sec:exp}
\begin{table}[t]
    \centering
    \resizebox{\linewidth}{!}{
    \begin{tabular}{lccccc}
        \toprule
        \textbf{Method} & \makecell{\textbf{ViT-}\\ \textbf{based}} & \makecell{\textbf{CNN-}\\ \textbf{based}} & \makecell{\textbf{Diverse-Origin}\\ \textbf{Models}}  & \textbf{Ensemble} & \makecell{\textbf{Theoratical}\\ \textbf{Support}}  \\ 
        \midrule
        Simple-Averaging\cite{wortsman2022model}  & \cmark & \cmark$^{*}$ & \xmark & \cmark & \cmark$^{*}$ \\
        Task-Arithmetic\cite{ilharcoediting}  & \cmark & \cmark$^{*}$ & \xmark & \cmark & \xmark \\
        Ties-Merging\cite{yadav2024ties}  & \xmark & \xmark & \xmark & \xmark & \xmark \\
        RegMean\cite{jin2023dataless}   & \xmark & \xmark & \xmark & \xmark & \xmark \\
        AdaMerging\cite{yangadamerging}  & \cmark & \xmark & \xmark & \cmark & \xmark \\
        WeMoE\cite{tangmerging}  & \cmark & \xmark & \xmark & \cmark & \xmark \\
        \midrule
        \textbf{\texttt{NeuLig} (Ours)} & \cmark & \cmark & \cmark & \cmark & \cmark \\
        \bottomrule
    \end{tabular}
    }
    \caption{The asterisk indicates that the condition is `partially satisfied'. For Simple-Averaging, the theoretical discussion is limited to the relationship between the performance of merging two models and that of ensembling\cite{wortsman2022model}. Furthermore, although both Simple-Averaging and Task-Arithmetic can be applied to CNN-based models, their performance is suboptimal. In the case of Diverse-Origin Models, all previous methods yield performance close to random guessing, but our conclusions remain applicable.
    }
    \label{compare}
    \vspace{-3mm}
\end{table}
\subsection{Models and Datasets}
\label{mo_da}
In this work, we consider CLIP-like models, which is in line with common practice \cite{ortiz2023task,yangadamerging,yadav2024ties}. We use two types of pre-trained models: CLIP-RN50 and CLIP-ViT-B/32 from OpenCLIP \cite{ilharco_gabriel_2021_5143773}, and each is fine-tuned on five datasets: GTSRB, CIFAR10, RESISC45, CIFAR100, and MNIST. Following the previous model merging setup, we freeze the text encoder of each model and focus solely on the visual component, ensuring that each class in the dataset has an identical language feature representation across models to prevent feature space collapse and conflicts \cite{ilharcoediting}. 

\begin{table*}[ht]
    \centering
    \scriptsize
    \renewcommand{\arraystretch}{1.3}
    \setlength{\tabcolsep}{3pt}
    
    \begin{subtable}[t]{\textwidth}
        \centering
        \resizebox{\textwidth}{!}{
            \begin{tabular}{@{}lcccccccccccccccccc@{}}
                \toprule
                \textbf{Method} & 
                \multicolumn{3}{c}{\textbf{GTSRB}} & 
                \multicolumn{3}{c}{\textbf{CIFAR100}} & 
                \multicolumn{3}{c}{\textbf{RESISC45}} & 
                \multicolumn{3}{c}{\textbf{CIFAR10}} & 
                \multicolumn{3}{c}{\textbf{MNIST}} & 
                \multicolumn{3}{c}{\textbf{Avg}} \\
                \midrule
                Pre-trained & \multicolumn{3}{c}{35.06} & \multicolumn{3}{c}{40.30} & \multicolumn{3}{c}{54.35} & \multicolumn{3}{c}{71.57} & \multicolumn{3}{c}{57.60} & \multicolumn{3}{c}{51.78} \\
                Fine-tuned & \multicolumn{3}{c}{97.89} & \multicolumn{3}{c}{76.69} & \multicolumn{3}{c}{91.71} & \multicolumn{3}{c}{93.72} & \multicolumn{3}{c}{99.56} & \multicolumn{3}{c}{91.91} \\
                \midrule
                & \textbf{Mer. $\uparrow$} & \textbf{Ens. $\uparrow$} & \textbf{Gap $ \downarrow $} &
                \textbf{Mer. $\uparrow$} & \textbf{Ens. $\uparrow$} & \textbf{Gap $ \downarrow $} & 
                \textbf{Mer. $\uparrow$} & \textbf{Ens. $\uparrow$} & \textbf{Gap $ \downarrow $} & 
                \textbf{Mer. $\uparrow$} & \textbf{Ens. $\uparrow$} & \textbf{Gap $ \downarrow $} & 
                \textbf{Mer. $\uparrow$} & \textbf{Ens. $\uparrow$} & \textbf{Gap $ \downarrow $} & 
                \textbf{Mer. $\uparrow$} & \textbf{Ens. $\uparrow$} & \textbf{Gap $ \downarrow $}  \\
                \cmidrule(lr){2-4} \cmidrule(lr){5-7} 
                \cmidrule(lr){8-10} \cmidrule(lr){11-13} \cmidrule(lr){14-16} \cmidrule(lr){17-19}
                &&&&&&&\multicolumn{6}{l}{\textit{Multi-Task Model Collaboration Methods}}&&&& \\
                Simple-Averaging\cite{wortsman2022model} & 52.29 & 99.81 & 47.52 & 34.27 & 62.84 & 28.57 & 18.89 & 91.63 & 72.74 & 94.82 & 76.81 & 18.01 & 99.52 & 69.70 & 29.82 & 50.39 & 89.73 & 39.34 \\
                Task-Arithmetic\cite{ilharcoediting} & 47.19 & 99.80 & 52.61 & 41.44 & 62.58 & 21.14 & 37.30 & 91.43 & 54.13 & 76.98 & 94.97 & 17.99 & 64.58 & 99.56 & 34.98 & 53.50 & 89.67 & 36.17 \\
                Ties-Merging\cite{yadav2024ties} & 43.53 & - & - & 28.98 & - & - & 28.63 & - & - & 60.99 & - & - & 58.52 & - & - & 44.13 & - & -\\
                \midrule
                &&&&&&&&\multicolumn{3}{l}{\textit{Neural Ligand}}&&&&&& \\
                \cellcolor{teal!10}Ours (Semi-Supervised) & \cellcolor{teal!10}98.49 & \cellcolor{teal!10}98.66 & \cellcolor{orange!10}\textbf{0.17} & \cellcolor{teal!10}\textbf{78.92} & \cellcolor{teal!10}\textbf{79.31} & \cellcolor{orange!10}0.39 & \cellcolor{teal!10}\textbf{92.89} & \cellcolor{teal!10}93.11 & \cellcolor{orange!10}\textbf{1.97} & \cellcolor{teal!10}90.62 & \cellcolor{teal!10}94.60 & \cellcolor{orange!10}3.98 & \cellcolor{teal!10}99.44 & \cellcolor{teal!10}99.52 & \cellcolor{orange!10}0.08 & \cellcolor{teal!10}\textbf{92.07} \textcolor{teal}{(+38.57)} & \cellcolor{teal!10}92.68 \textcolor{teal}{(+2.95)} & \cellcolor{orange!10}\textbf{0.61} \textcolor{orange}{(-35.56)}\\
                \cellcolor{teal!10}Ours (Supervised) & \cellcolor{teal!10}\textbf{99.26} & \cellcolor{teal!10}\textbf{99.82} & \cellcolor{orange!10}0.56 & \cellcolor{teal!10}77.00 & \cellcolor{teal!10}{77.01} & \cellcolor{orange!10}\textbf{0.01} & \cellcolor{teal!10}88.21 & \cellcolor{teal!10}\textbf{93.17} & \cellcolor{orange!10}4.96 & \cellcolor{teal!10}\textbf{92.41} & \cellcolor{teal!10}\textbf{94.89} & \cellcolor{orange!10}\textbf{2.48} & \cellcolor{teal!10}\textbf{99.49} & \cellcolor{teal!10}\textbf{99.55} & \cellcolor{orange!10}\textbf{0.06} & \cellcolor{teal!10}91.47 \textcolor{teal}{(+37.97)} & \cellcolor{teal!10}\textbf{92.69} \textcolor{teal}{(+2.96)} & \cellcolor{orange!10}1.22 \textcolor{orange}{(-34.95)} \\
                \bottomrule
            \end{tabular}
        }
        \caption{Results of different methods on various datasets using CLIP-RN50 (RegMean, AdaMerging, and WeMoE are not applicable to ResNet-based models).}
        \label{tab:main_rn50}
    \end{subtable}
    
    \vspace{1em}
    
    \begin{subtable}[t]{\textwidth}
        \centering
        \resizebox{\textwidth}{!}{
            \begin{tabular}{lcccccccccccccccccc}
                \toprule
                \textbf{Method} & 
                \multicolumn{3}{c}{\textbf{GTSRB}} & 
                \multicolumn{3}{c}{\textbf{CIFAR100}} & 
                \multicolumn{3}{c}{\textbf{RESISC45}} & 
                \multicolumn{3}{c}{\textbf{CIFAR10}} & 
                \multicolumn{3}{c}{\textbf{MNIST}} & 
                \multicolumn{3}{c}{\textbf{Avg}} \\
                \midrule
                Pre-trained & \multicolumn{3}{c}{32.56} & \multicolumn{3}{c}{64.20} & \multicolumn{3}{c}{60.22} & \multicolumn{3}{c}{89.83} & \multicolumn{3}{c}{48.25} & \multicolumn{3}{c}{59.01} \\
                Fine-tuned & \multicolumn{3}{c}{98.95} & \multicolumn{3}{c}{84.22} & \multicolumn{3}{c}{94.13} & \multicolumn{3}{c}{97.13} & \multicolumn{3}{c}{99.56} & \multicolumn{3}{c}{94.80} \\
                \midrule
                & \textbf{Mer. $\uparrow$} & \textbf{Ens. $\uparrow$} & \textbf{Gap $ \downarrow $} &
                \textbf{Mer. $\uparrow$} & \textbf{Ens. $\uparrow$} & \textbf{Gap $ \downarrow $} & 
                \textbf{Mer. $\uparrow$} & \textbf{Ens. $\uparrow$} & \textbf{Gap $ \downarrow $} & 
                \textbf{Mer. $\uparrow$} & \textbf{Ens. $\uparrow$} & \textbf{Gap $ \downarrow $} & 
                \textbf{Mer. $\uparrow$} & \textbf{Ens. $\uparrow$} & \textbf{Gap $ \downarrow $} & 
                \textbf{Mer. $\uparrow$} & \textbf{Ens. $\uparrow$} & \textbf{Gap $ \downarrow $}  \\
                \cmidrule(lr){2-4} \cmidrule(lr){5-7} 
                \cmidrule(lr){8-10} \cmidrule(lr){11-13} \cmidrule(lr){14-16} \cmidrule(lr){17-19}
                &&&&&&&\multicolumn{6}{l}{\textit{Multi-Task Model Collaboration Methods}}&&&& \\
                Simple-Averaging\cite{wortsman2022model} & 59.30 & 92.16 & 32.86 & 75.46 & 78.13 & 2.67 & 73.17 & 84.25 & 11.08 & 95.37 & 97.21 & 1.84 & 87.65 & 98.82 & 11.17 & 78.19 & 90.11 & 11.92 \\
                Task-Arithmetic\cite{ilharcoediting} & 64.49 & 92.12 & 27.63 & 73.38 & 78.17 & 4.79 & 69.11 & 84.08 & 14.97 & 94.90 & 97.23 & 2.33 & 91.51 & 98.80 & 7.29 & 78.68 & 90.08 & 11.40\\
                Ties-Merging\cite{yadav2024ties} & 64.40 & - & - & 76.17 & - & - & 76.16 & - & - & 95.82 & - & - & 91.46 & - & - & 80.80 & - & -\\
                RegMean\cite{jin2023dataless} & 76.72 & - & - & 70.04 & - & - & 79.94 & - & - & 95.89 & - & - & 93.19 & - & - & 83.16 & - & - \\
                AdaMerging\cite{yangadamerging} & 91.49 & 92.83 & 1.34 & 75.17 & 76.19 & 1.02 & 84.32 & 89.67 & 5.35 & 94.90 & 96.72 & 1.82 & 95.87 & 97.88 & 1.52 & 88.65 & 90.25 & 1.60 \\
                WeMoE\cite{tangmerging} & 92.28 & 92.49 & 0.21 & 77.15 & 78.26 & 1.11 & 86.36 & 86.94 & 0.58 & 95.57 & 98.69 & 3.12 & 96.73 & 99.03 & 2.30 & 89.76 & 91.08 & 1.32 \\
                \midrule
                &&&&&&&&\multicolumn{3}{l}{\textit{Neural Ligand}}&&&&&& \\
                \cellcolor{teal!10}Ours (Semi-Supervised) & \cellcolor{teal!10}\textbf{99.83} & \cellcolor{teal!10}99.80 & \cellcolor{orange!10}\textbf{0.10} & \cellcolor{teal!10}\textbf{86.26} & \cellcolor{teal!10}\textbf{86.28} & \cellcolor{orange!10}\textbf{0.02} & \cellcolor{teal!10}\textbf{94.54} & \cellcolor{teal!10}\textbf{94.48} & \cellcolor{orange!10}0.64 & \cellcolor{teal!10}96.98 & \cellcolor{teal!10}\textbf{97.17} & \cellcolor{orange!10}0.77 & \cellcolor{teal!10}99.58 & \cellcolor{teal!10}99.58 & \cellcolor{orange!10}\textbf{0.00} & \cellcolor{teal!10}\textbf{95.44} \textcolor{teal}{(+5.68)} & \cellcolor{teal!10}\textbf{95.46} \textcolor{teal}{(+4.38)} & \cellcolor{orange!10}\textbf{0.02} \textcolor{orange}{(-1.30)} \\
                \cellcolor{teal!10}Ours (Supervised) & \cellcolor{teal!10}99.67 & \cellcolor{teal!10}\textbf{99.90} & \cellcolor{orange!10}0.23 & \cellcolor{teal!10}86.17 & \cellcolor{teal!10}86.00 & \cellcolor{orange!10}0.17 & \cellcolor{teal!10}94.40 & \cellcolor{teal!10}94.40 & \cellcolor{orange!10}\textbf{0.00} & \cellcolor{teal!10}\textbf{96.99} & \cellcolor{teal!10}96.60 & \cellcolor{orange!10}\textbf{0.39} & \cellcolor{teal!10}\textbf{99.60} & \cellcolor{teal!10}\textbf{99.60} & \cellcolor{orange!10}\textbf{0.00} & \cellcolor{teal!10}95.37 \textcolor{teal}{(+5.61)} & \cellcolor{teal!10}95.30 \textcolor{teal}{(+4.22)} & \cellcolor{orange!10}0.07 \textcolor{orange}{(-1.25)}\\
                \bottomrule
            \end{tabular}
        }
        \caption{Results of different methods on various datasets using CLIP-ViT-B/32.}
        \label{tab:main_vit}
        \vspace{-3mm}
    \end{subtable}
    \caption{Results of various methods across multiple datasets, including the merging performance, the ensembling performance, and the performance gap for both CLIP-RN50 and CLIP-ViT-B/32.}
    \vspace{-3mm}
    \label{tab:datalevel_main}
\end{table*}
\subsection{Baseline Methods}
Since \texttt{NeuLig} is a validation framework designed for exploring the properties of model merging, we use six popular model merging baselines for comparison, including the static merging methods Simple-Averaging \cite{wortsman2022model}, Task-Arithmetic \cite{ilharcoediting}, Ties-Merging \cite{yadav2024ties}, and RegMean \cite{jin2023dataless}, as well as the learnable merging methods AdaMerging \cite{yangadamerging} and WeMoE \cite{tangmerging}. Among these methods, as discussed in Table \ref{compare}, RegMean, AdaMerging, and WeMoE are not applicable to ResNet-based models. Ties-Merging and RegMean are incompatible with ensembling. All of them can not be used in the divers-origin model scenario and can merely yield performance close to random guessing. Results in Table \ref{tab:datalevel_main} clearly illustrate their limitations.

\subsection{NeuLig under Multi-Model Collaboration}
We first use \texttt{NeuLig} to verify whether the performance consistency can be achieved between merging and ensembling at the data level, i.e., we use Portland to generate a unique CoopVec for each data, which is then applied to merging and ensembling. Results when 5 models collaborating are shown in Table \ref{tab:datalevel_main}. We report the merging performance (Mer.), ensembling performance (Ens.), and performance gap for each method on each dataset, along with the average performance and average performance gap. We also report the performance of both the pre-trained model and each individual fine-tuned model. The best performances for merging, ensembling, and performance gap are highlighted in bold.

The results can be analyzed from several key perspectives. First, considering the performance gap, we observe that for baseline methods compatible with both merging and ensembling, ensembling typically outperforms merging, with a noticeable performance gap in most cases. This phenomenon is consistent with conclusions drawn in previous works \cite{wortsman2022model,wortsman2022robust,stoica2023zipit}. In terms of architecture-specific trends, baseline methods generally perform better in merging for ViT-based models compared to ResNet-based models. This may due to the modular, attention-driven architecture of ViTs, which facilitates task compatibility during merging. ViTs rely on self-attention mechanisms that produce globally generalized representations, making merged parameters more stable and transferable across tasks \cite{dosovitskiy2021an}. In contrast, ResNet’s convolutional structure depends heavily on localized spatial relationships, meaning that parameter-level merging can disrupt these critical feature connections \cite{he2016deep}. Moreover, model merging relies on the linear interpolative nature of task vectors, a property that aligns well with the weight space of ViTs but is less compatible with the structured convolutional filters in ResNet models \cite{wortsman2022model}. For learnable baseline methods like AdaMerging and WeMoE, their performance consistency exceeds that of static methods. This result aligns with intuition, as learnable methods are designed to adaptively adjust task vectors on a case-by-case basis (i.e., data level), allowing for a refined alignment between the merged models and target tasks \cite{rebuffi2018efficient}. 

In contrast, \texttt{NeuLig} reliably demonstrates high performance consistency and substantial utility across various scenarios, underscoring its effectiveness as a validation framework. Specifically, in the collaboration between ViT-based models, we observe a performance gap with precision to the second decimal place, whereas for ResNet-based models, the gap remains minimal at approximately \(1\%\). This further supports that merging can achieve data-level performance consistency comparable to ensembling.

\begin{figure*}[ht]
    \centering
    \begin{minipage}{\textwidth}
    \captionsetup[subfigure]{labelformat=empty} 
        \begin{subfigure}{\textwidth}
            \centering
            \begin{subfigure}{0.16\textwidth}
                \centering
                \includegraphics[width=\linewidth]{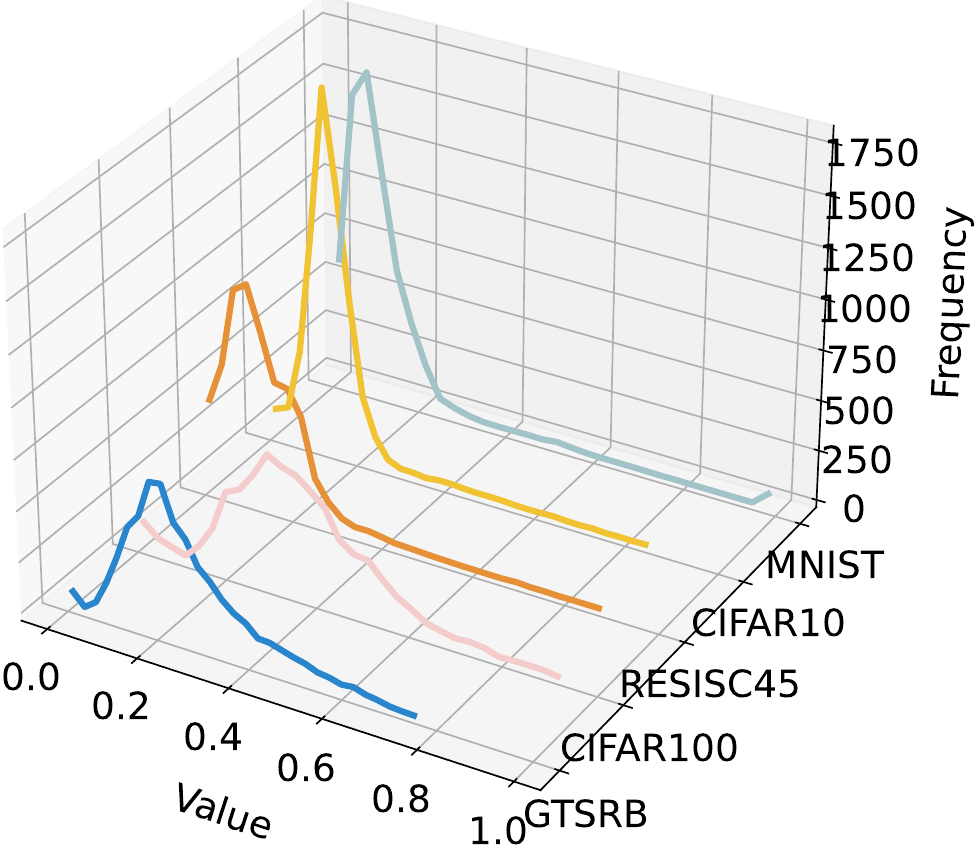} 
            \caption{GTSRB}
            \end{subfigure}
            \begin{subfigure}{0.16\textwidth}
                \centering
                \includegraphics[width=\linewidth]{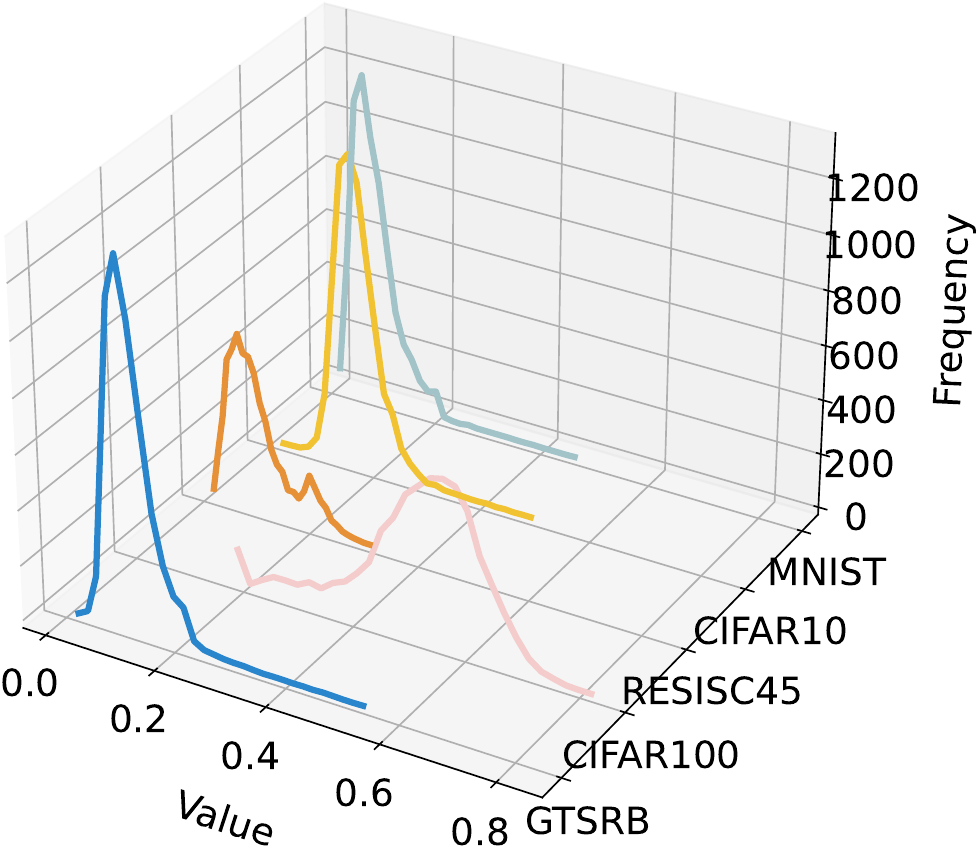} 
            \caption{CIFAR100}
            \end{subfigure}
            \begin{subfigure}{0.16\textwidth}
                \centering
                \includegraphics[width=\linewidth]{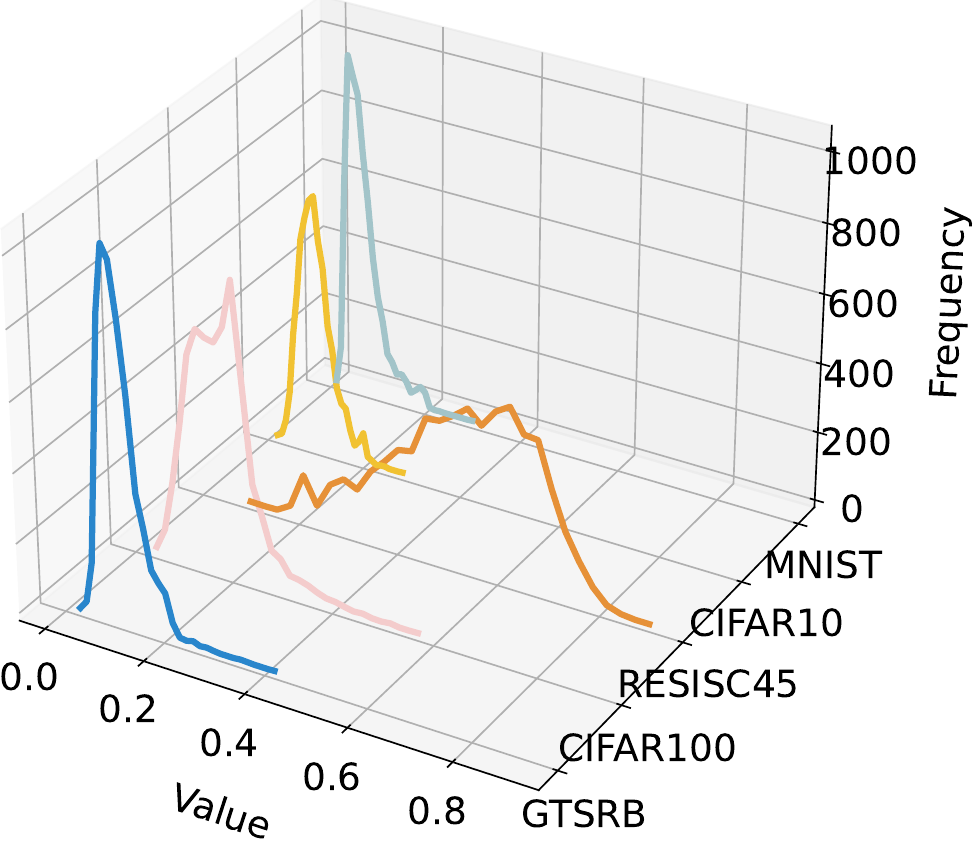}
                \caption{RESISC45}
            \end{subfigure}
            \begin{subfigure}{0.16\textwidth}
                \centering
                \includegraphics[width=\linewidth]{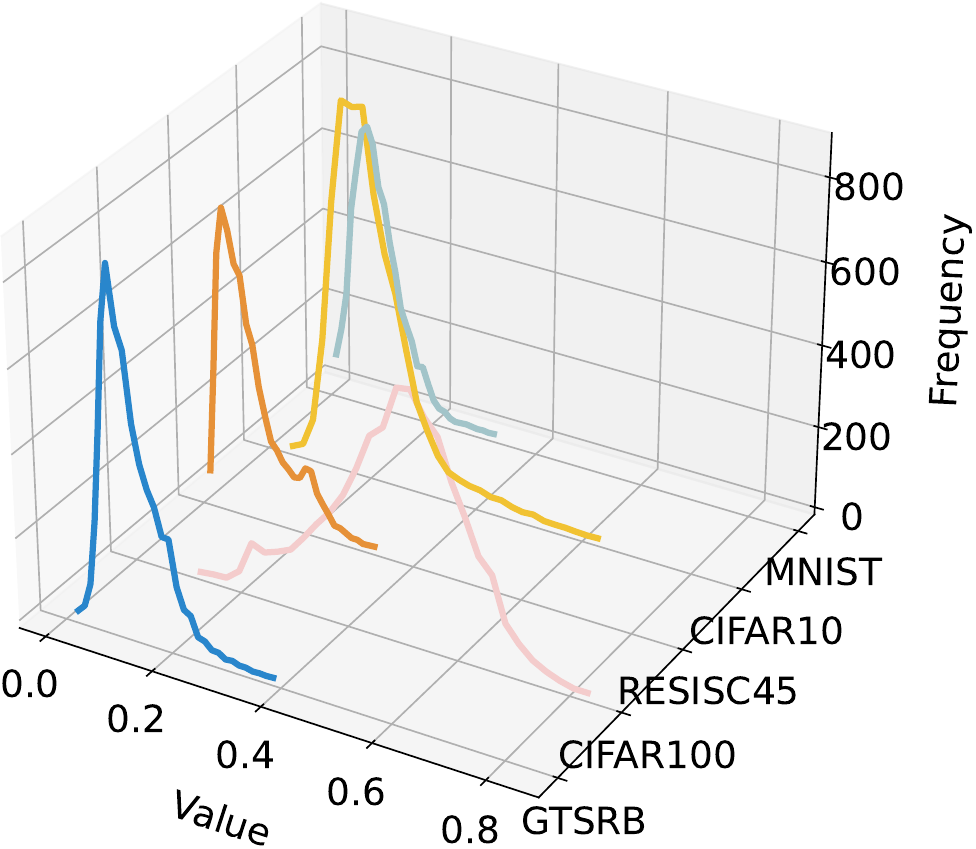}
                \caption{CIFAR10}
            \end{subfigure}
            \begin{subfigure}{0.16\textwidth}
                \centering
                \includegraphics[width=\linewidth]{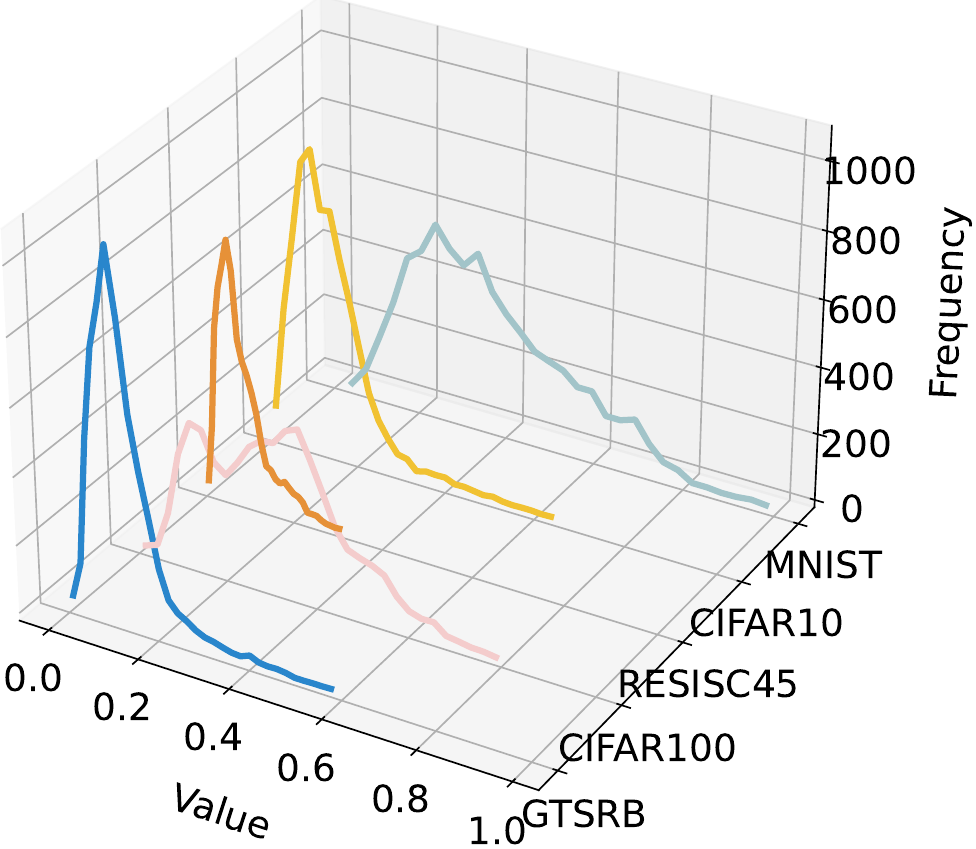}
                \caption{MNIST}
            \end{subfigure}
            \begin{subfigure}{0.16\textwidth}
                \centering
                \includegraphics[width=\linewidth]{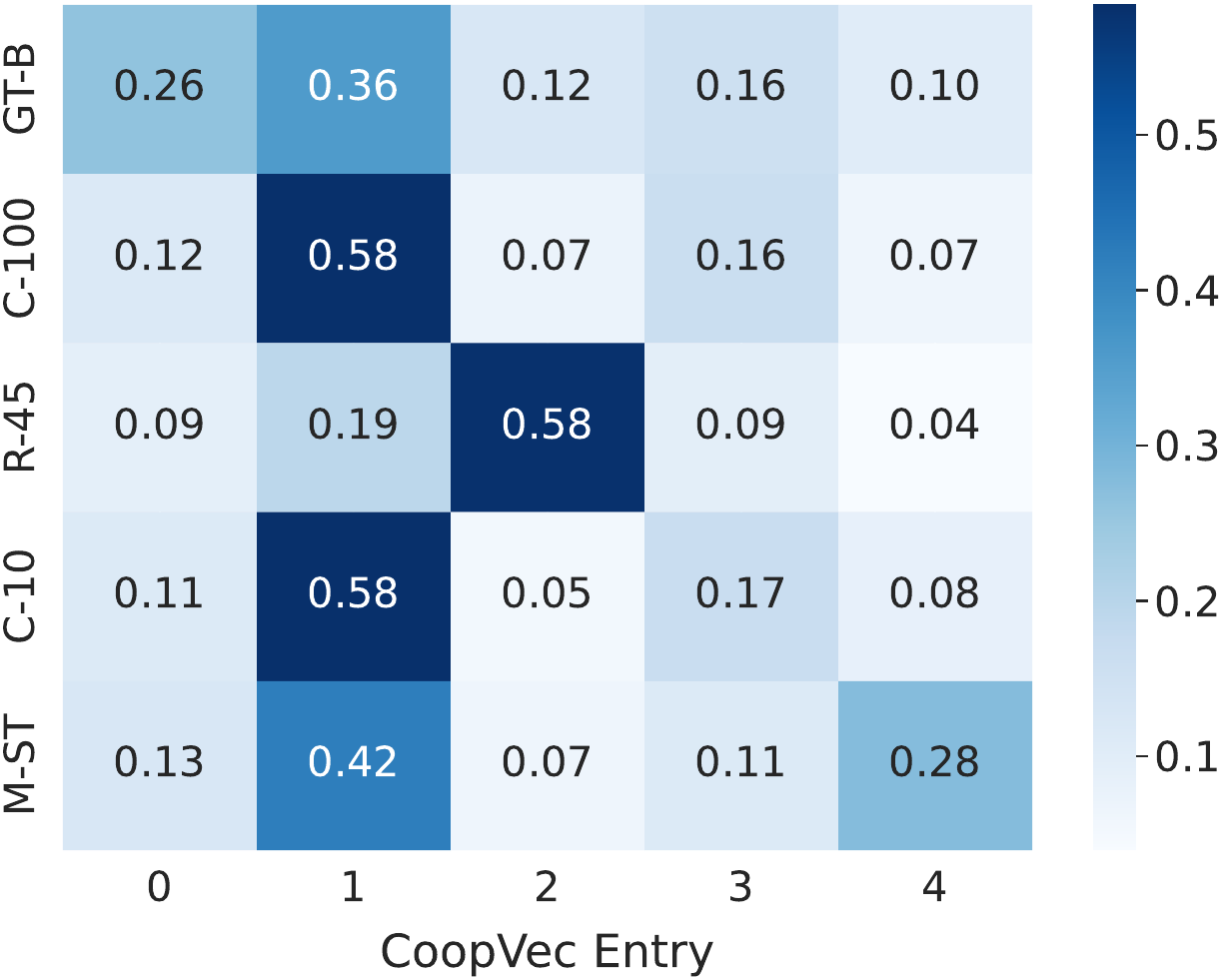} 
                \caption{CoopVec Map}
            \end{subfigure}
            \caption{(a) CLIP-RN50.}
        \end{subfigure}
        \begin{subfigure}{\textwidth}
            \centering
            \begin{subfigure}{0.16\textwidth}
                \centering
                \includegraphics[width=\linewidth]{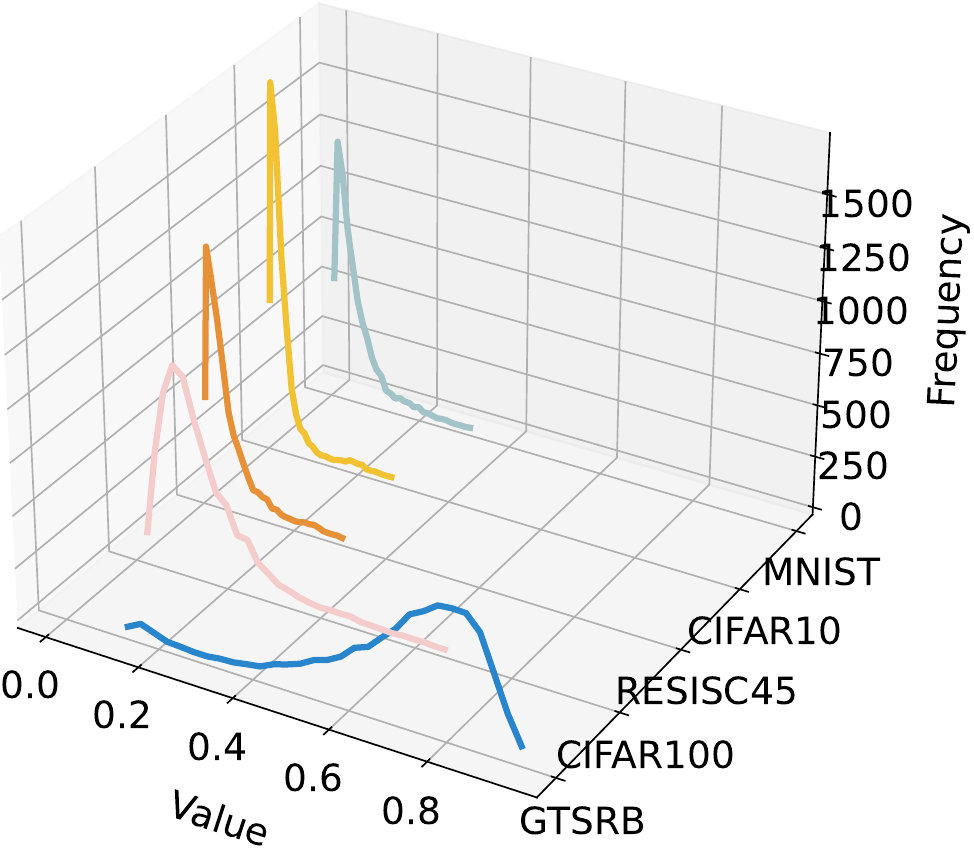} 
                \caption{GTSRB}
            \end{subfigure}
            \begin{subfigure}{0.16\textwidth}
                \centering
                \includegraphics[width=\linewidth]{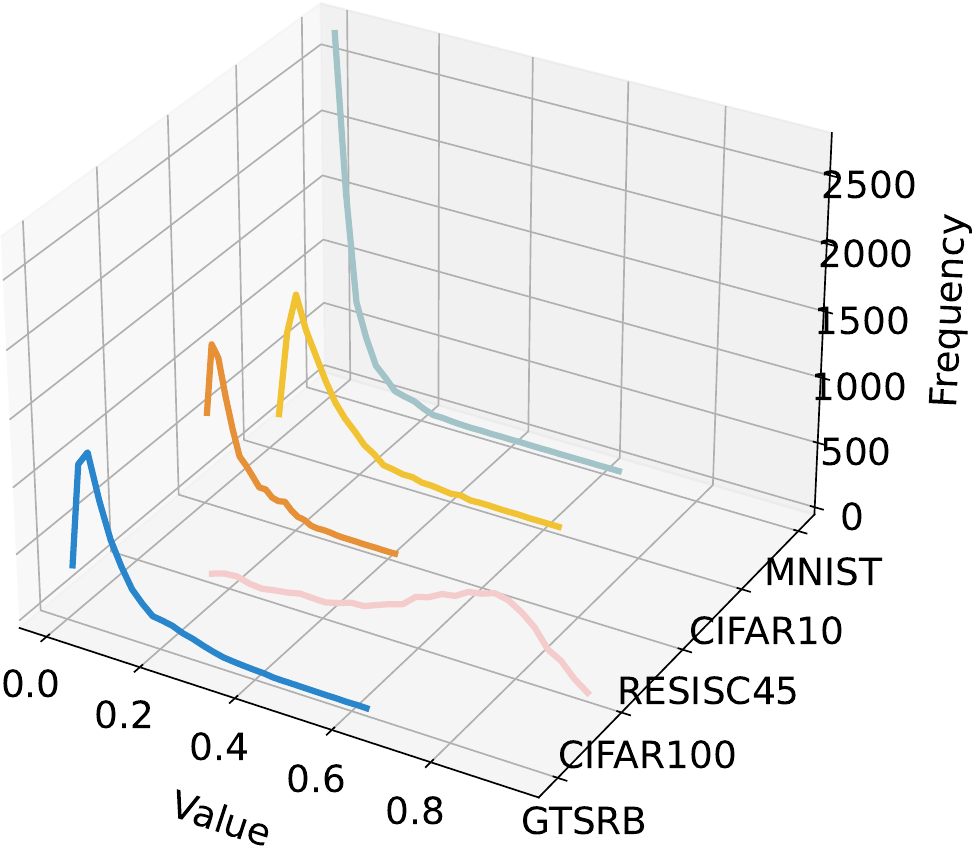} 
                \caption{CIFAR100}
            \end{subfigure}
            \begin{subfigure}{0.16\textwidth}
                \centering
                \includegraphics[width=\linewidth]{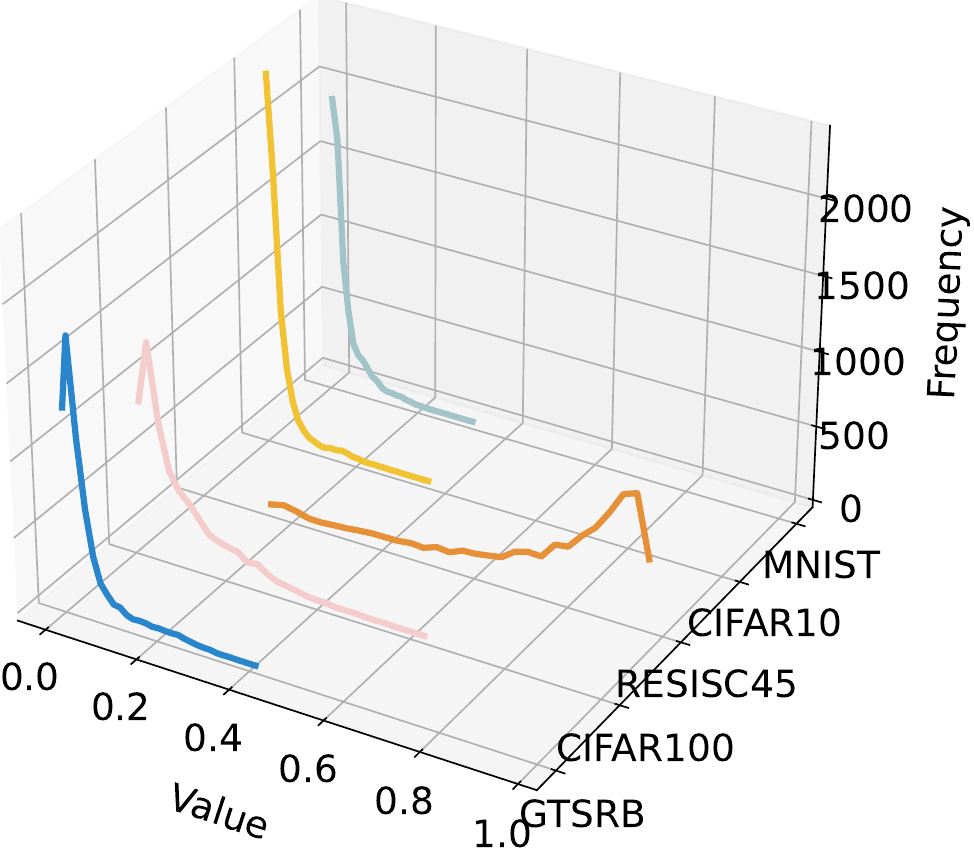}
                \caption{RESISC45}
            \end{subfigure}
            \begin{subfigure}{0.16\textwidth}
                \centering
                \includegraphics[width=\linewidth]{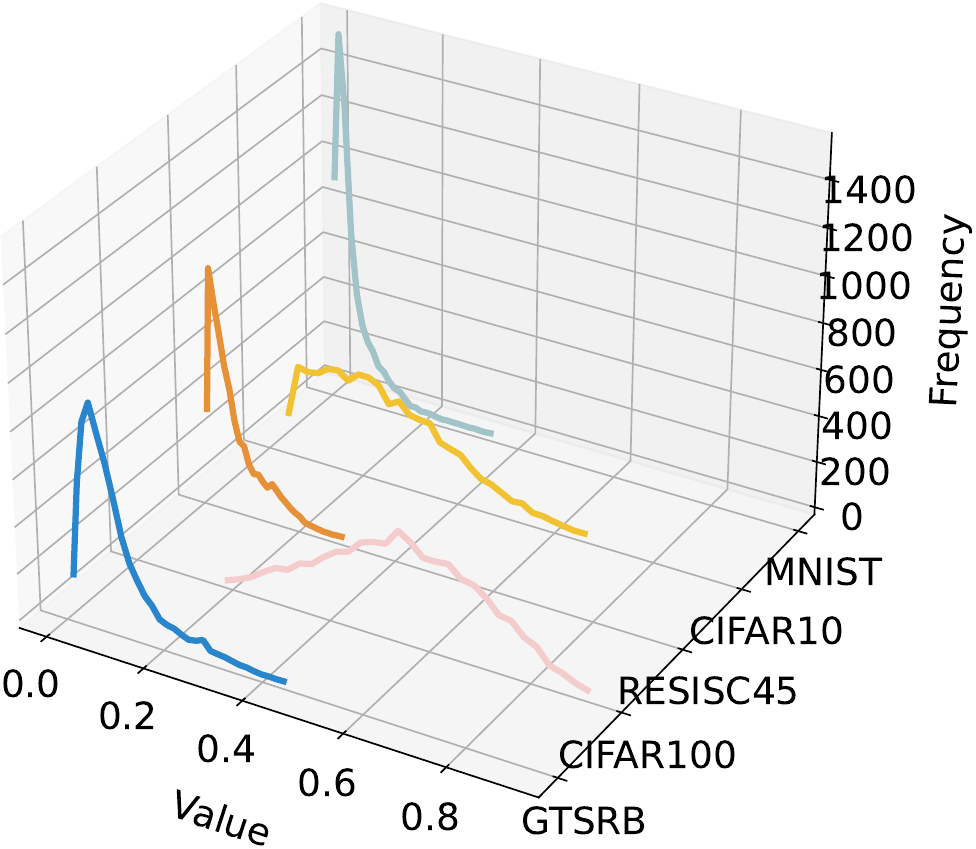}
                \caption{CIFAR10}
            \end{subfigure}
            \begin{subfigure}{0.16\textwidth}
                \centering
                \includegraphics[width=\linewidth]{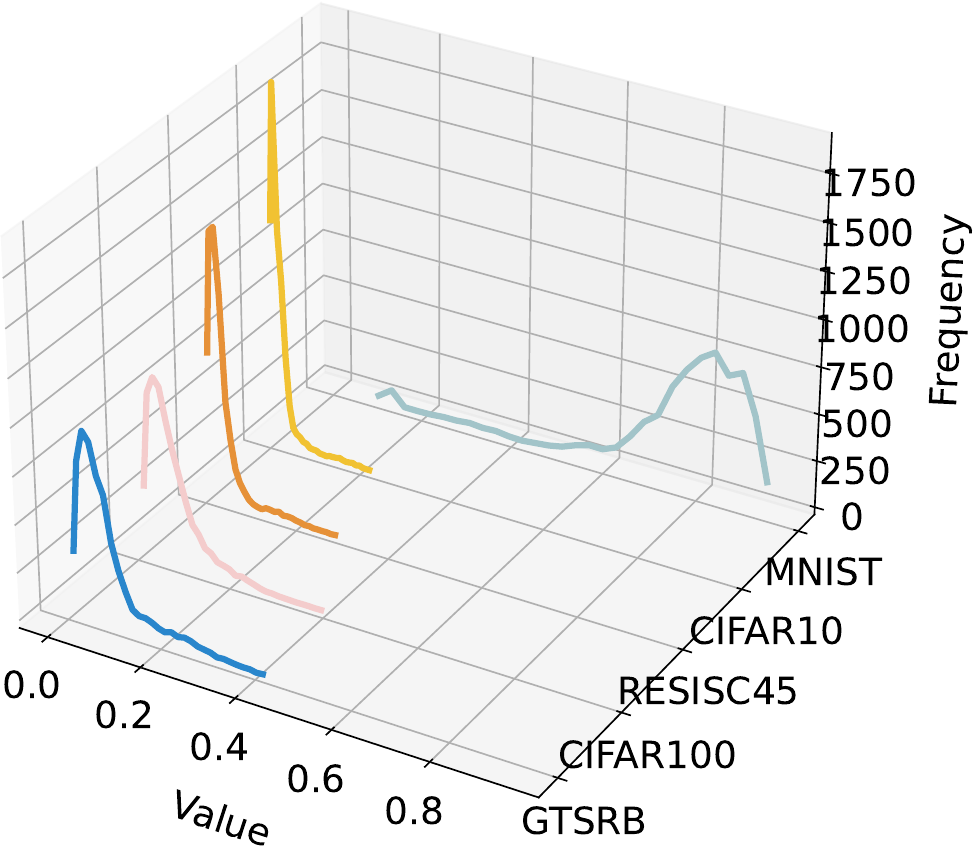}
                \caption{MNIST}
            \end{subfigure}
            \begin{subfigure}{0.16\textwidth}
                \centering
                \includegraphics[width=\linewidth]{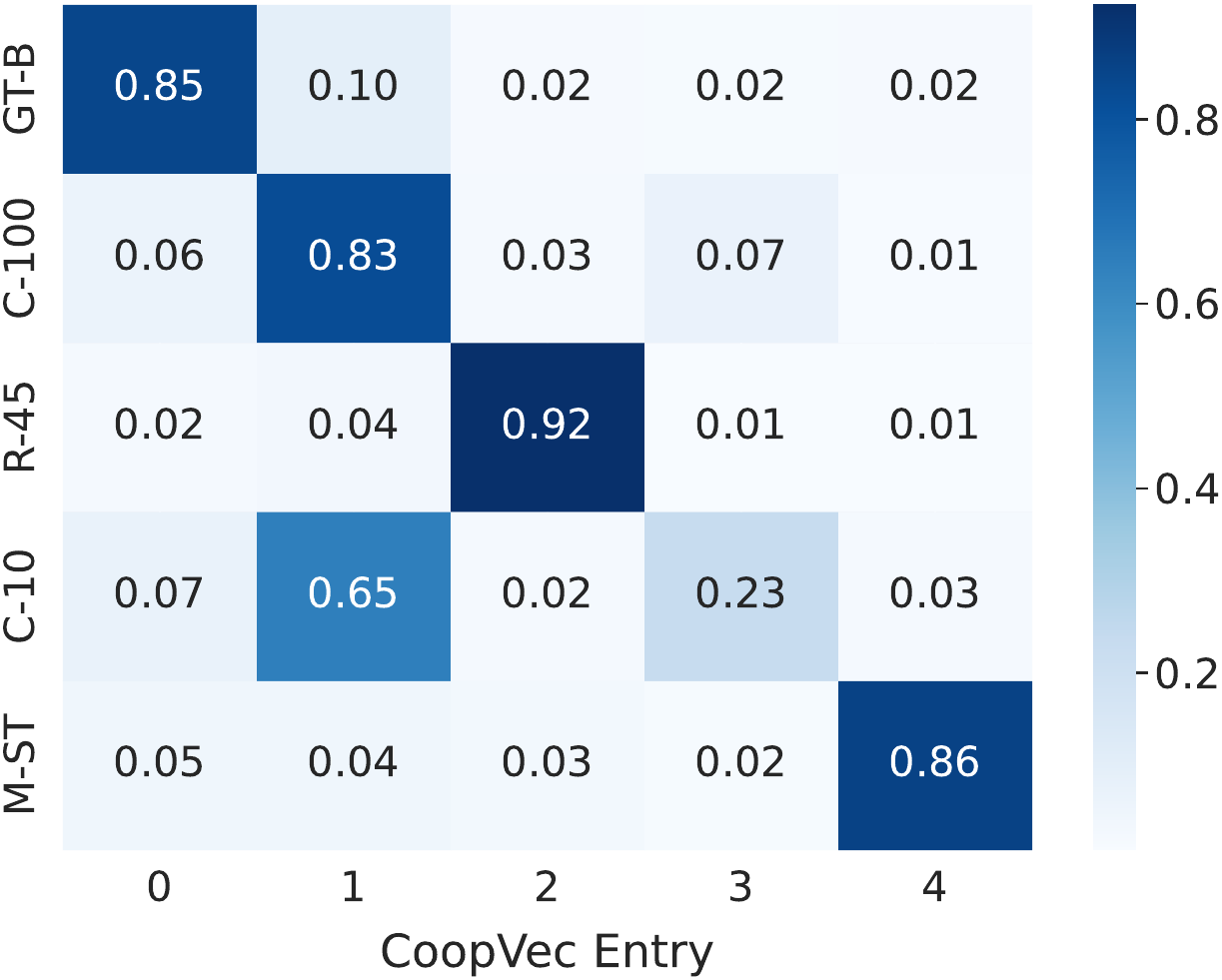}
                \caption{CoopVec Map}
            \end{subfigure}
            \caption{(b) CLIP-ViT-B/32.}
        \end{subfigure}
        \vspace{-7mm}
    \end{minipage}
    \caption{CoopVec Distribution of different tasks and the corresponding CoopVec Map after training for one epoch.}
    \label{fig:coopvec_map}
    \vspace{-5mm}
\end{figure*}
\begin{figure}[h]
    \centering
    \begin{minipage}{0.51\linewidth} 
        \centering
        \includegraphics[width=0.95\linewidth]{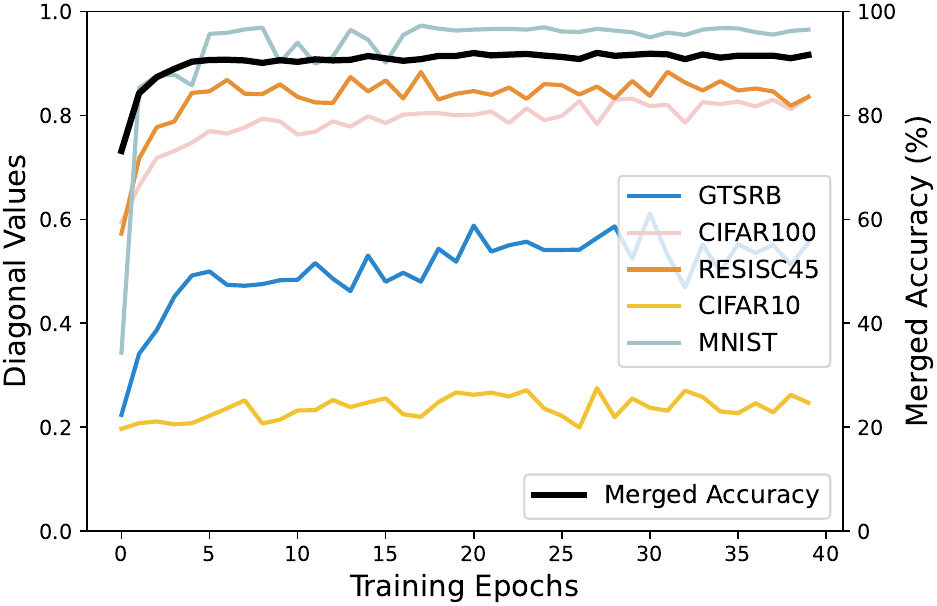}
        \includegraphics[width=0.95\linewidth]{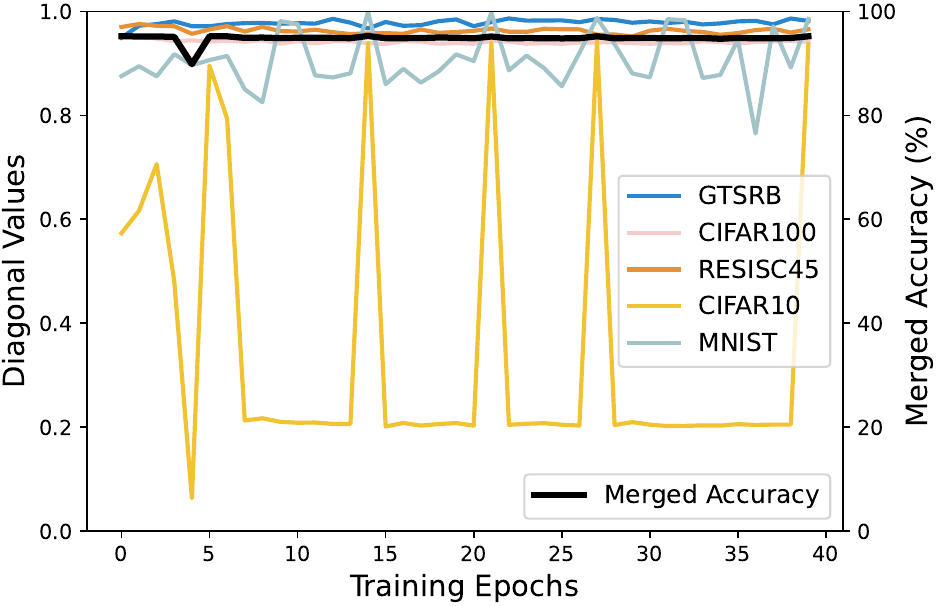}
        \caption{The variation of the diagonal values of CoopVec Map throughout the training process using CLIP-RN50 (top) and CLIP-ViT-B/32 (bottom).}
        \label{map_train_curve}
        \vspace{-5mm}
    \end{minipage}%
    \hspace{0.03\linewidth} 
    \begin{minipage}{0.45\linewidth} 
        \renewcommand{\arraystretch}{1.8} 
        \setlength{\tabcolsep}{4pt}  
        \centering
        \captionof{table}{The final merging (Mer.) performance and ensembling (Ens.) performance when using the CoopVec Map for collaboration.}
        \begin{adjustbox}{width=\linewidth} 
            \begin{tabular}{c|c|c|c|c}
                \hline
                \multirow{2}{*}{\textbf{Dataset}} & \multicolumn{4}{c}{\textbf{Performance}} \\ \cline{2-5}
                & \textbf{Mer.}\textcolor{pink}{\ding{117}} & \textbf{Ens.}\textcolor{teal}{\ding{108}} & \textbf{Mer.}\textcolor{pink}{\ding{117}} & \textbf{Ens.}\textcolor{teal}{\ding{108}} \\ \hline
                \multirow{2}{*}{GTSRB} & \multicolumn{2}{c|}{CLIP-RN50} & \multicolumn{2}{c}{CLIP-ViT-B/32} \\ \cline{2-5}
                & 97.95 & 97.70 & 98.85 & 98.92 \\ \hline
                \multirow{2}{*}{CIFAR100} & \multicolumn{2}{c|}{CLIP-RN50} & \multicolumn{2}{c}{CLIP-ViT-B/32} \\ \cline{2-5}
                & 77.53 & 77.30 & 86.2 & 86.02 \\ \hline
                \multirow{2}{*}{RESISC45} & \multicolumn{2}{c|}{CLIP-RN50} & \multicolumn{2}{c}{CLIP-ViT-B/32} \\ \cline{2-5}
                & 90.57 & 92.35 & 93.92 & 93.83 \\ \hline
                \multirow{2}{*}{CIFAR10} & \multicolumn{2}{c|}{CLIP-RN50} & \multicolumn{2}{c}{CLIP-ViT-B/32} \\ \cline{2-5}
                & 90.34 & 93.20 & 95.57 & 96.38 \\ \hline
                \multirow{2}{*}{MNIST} & \multicolumn{2}{c|}{CLIP-RN50} & \multicolumn{2}{c}{CLIP-ViT-B/32} \\ \cline{2-5}
                & 99.62 & 99.64 & 99.60 & 99.57 \\ \hline
                \multirow{1}{*}{Avg.Acc} & \makecell{91.20\\ \textcolor{teal}{(+37.70)}} & \makecell{92.04\\ \textcolor{teal}{(+2.31)}} & \makecell{94.83\\ \textcolor{teal}{(+5.07)}} & \makecell{94.94\\ \textcolor{teal}{(+3.86)}} \\ \hline
            \end{tabular}
            \label{map_table_perf}
        \end{adjustbox}
        \vspace{-5mm}
    \end{minipage}
\end{figure}
\subsection{CoopVec Map}
\label{coop_map}
The concept of the CoopVec Map is based on our observations in Figure \ref{fig:coopvec_map}, where we show the distributions of CoopVecs generated by Portland when processing inputs from different datasets after training for one epoch. Specifically, the x-axis represents the values of CoopVec entries, the y-axis denotes different entries, and the z-axis indicates the frequency. We can observe that all distributions exhibit a distinct peak, which is regarded as the key point that most significantly influences performance during collaboration. 

By extracting the peak value from each subfigure in Figure \ref{fig:coopvec_map} (each subfigure corresponds to a dataset), we obtain a vector with \(n\) entries, where \(n\) represents the number of collaborating models. Forming these vectors as a matrix results in an \(n \times n\) matrix. We then use the corresponding row in this matrix to collaborate on data from different datasets, thereby achieving dataset-level merging/ensembling. The visualization of the CoopVec Map after training Portland for one epoch is shown on the right side of Figure \ref{fig:coopvec_map}. 

We can analyze the results from several perspectives. Firstly, Figure \ref{fig:coopvec_map} indicates that the peak value in the distribution is already present at the early stage of training. For the ResNet-based models, CoopVec tends to assign higher weights to models fine-tuned on more complex, knowledge-rich datasets, such as CIFAR100 in our experiments. In contrast, for the ViT-based models, CoopVec prioritizes the model corresponding to each specific dataset, leading to a strongly orthogonal CoopVec Map from the outset. The greatest shifts from orthogonality are observed when tasks are similar, such as CIFAR10 and CIFAR100. Similar to the behavior observed in ResNet-based models, this shift tends to favor more complex datasets. In Figure \ref{map_train_curve}, we further show the variation in the diagonal values of the CoopVec Map throughout the training process. The final performance is reported in Table \ref{map_table_perf}. It can be observed that for CLIP-RN50, the diagonal values ultimately converge to positions that do not exhibit strong orthogonality (i.e., all values equal to 1). In contrast, for CLIP-ViT-B/32, all datasets except CIFAR10—which is highly correlated with CIFAR100—exhibit relatively high diagonal values, indicating stronger orthogonality. The values for CIFAR10 consistently display fluctuations, sometimes stabilizing at lower values with a strong shift toward CIFAR100, and at other times peaking sharply, suggesting strong orthogonality relative to other datasets. However, despite these fluctuations, the performance remains nearly stable. These phenomena highlight the different properties of ResNet-based and ViT-based models in multi-model collaboration. The former shows a stronger dependency on specific models, while the latter exhibits a tendency toward strong orthogonality in CoopVecs during collaboration. 

\subsection{Resilience of NeuLig}
\label{sec:experiments}

In this section, we conduct several ablation studies on \texttt{NeuLig}, including the training trajectory under different number of models, the performance under diverse-origin models, the computational resource consumption, and whether the performance consistency still exits when only a small subset of the data is available.

\begin{figure*}[ht]
  \centering
  \begin{subfigure}{0.29\linewidth}
    \centering
    \renewcommand{\arraystretch}{1.03} 
    \small
    \begin{tabular}{c|c|c|c}
      \hline
      \textbf{\# Models} & \multicolumn{3}{c}{\textbf{Performance}} \\ \hline
      \multirow{2}{*}{2} & \textcolor{teal}{\ding{108}}Ens.& \textcolor{pink}{\ding{117}}Mer. & Gap\\ \cline{2-4}
      & 88.32 & 88.15 & 0.17\\ \hline
      \multirow{2}{*}{3} & \textcolor{teal}{\ding{108}}Ens.& \textcolor{pink}{\ding{117}}Mer. & Gap \\ \cline{2-4}
      & 89.86 & 90.13 & 0.27\\ \hline
      \multirow{2}{*}{5} & \textcolor{teal}{\ding{108}}Ens.& \textcolor{pink}{\ding{117}}Mer. & Gap \\ \cline{2-4}
      & 92.69 & 91.47 & 1.22\\ \hline
    \end{tabular}
    \caption{Final Performance.}
    \label{fig:trajec}
  \end{subfigure}
  \begin{subfigure}{0.23\linewidth}
    \centering
    \includegraphics[width=\linewidth]{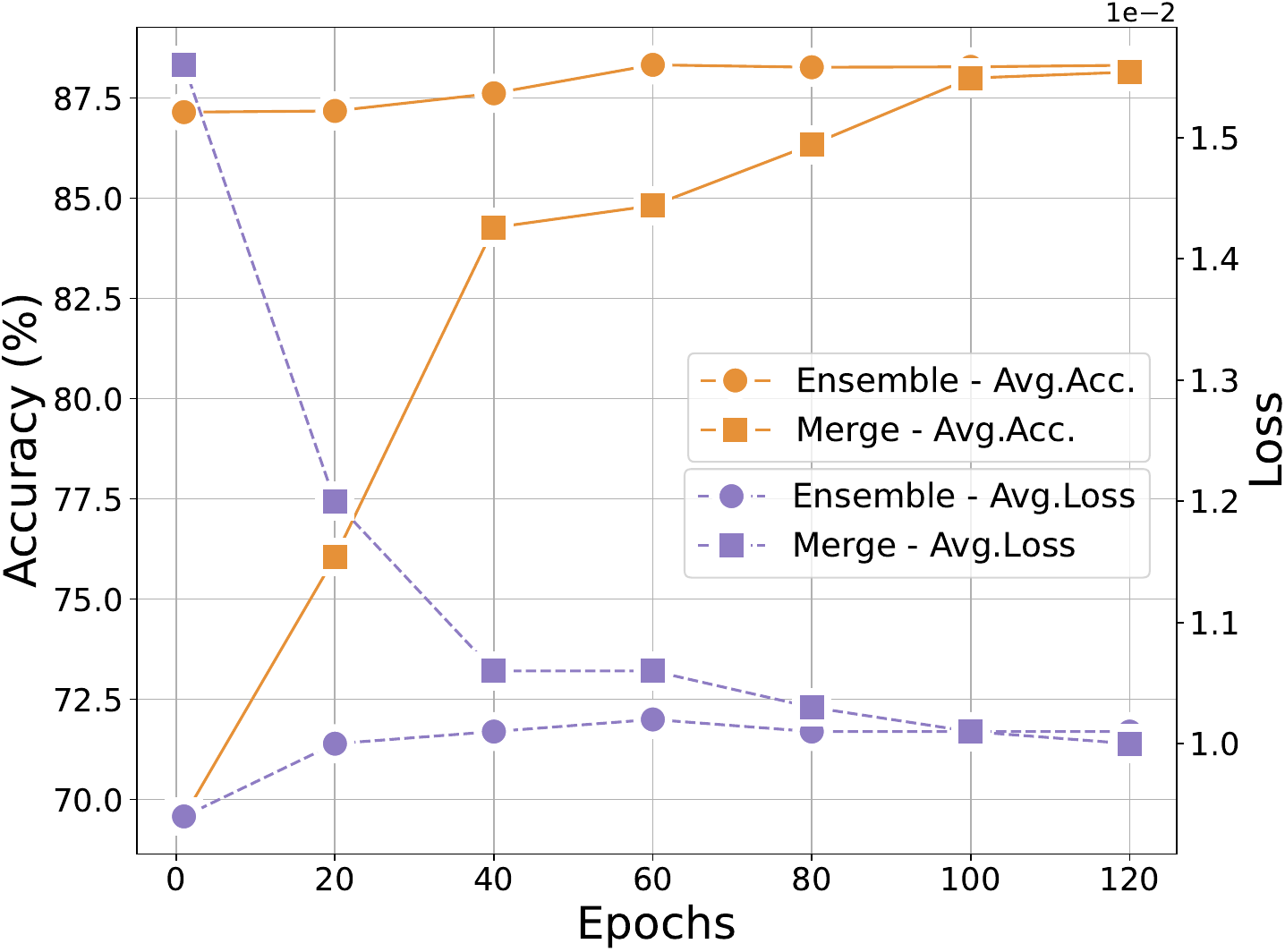}
    \caption{CLIP-RN50 (2 models).}
    \label{fig:short-2}
  \end{subfigure}
  \hfill
  \begin{subfigure}{0.225\linewidth}
    \centering
    \includegraphics[width=\linewidth]{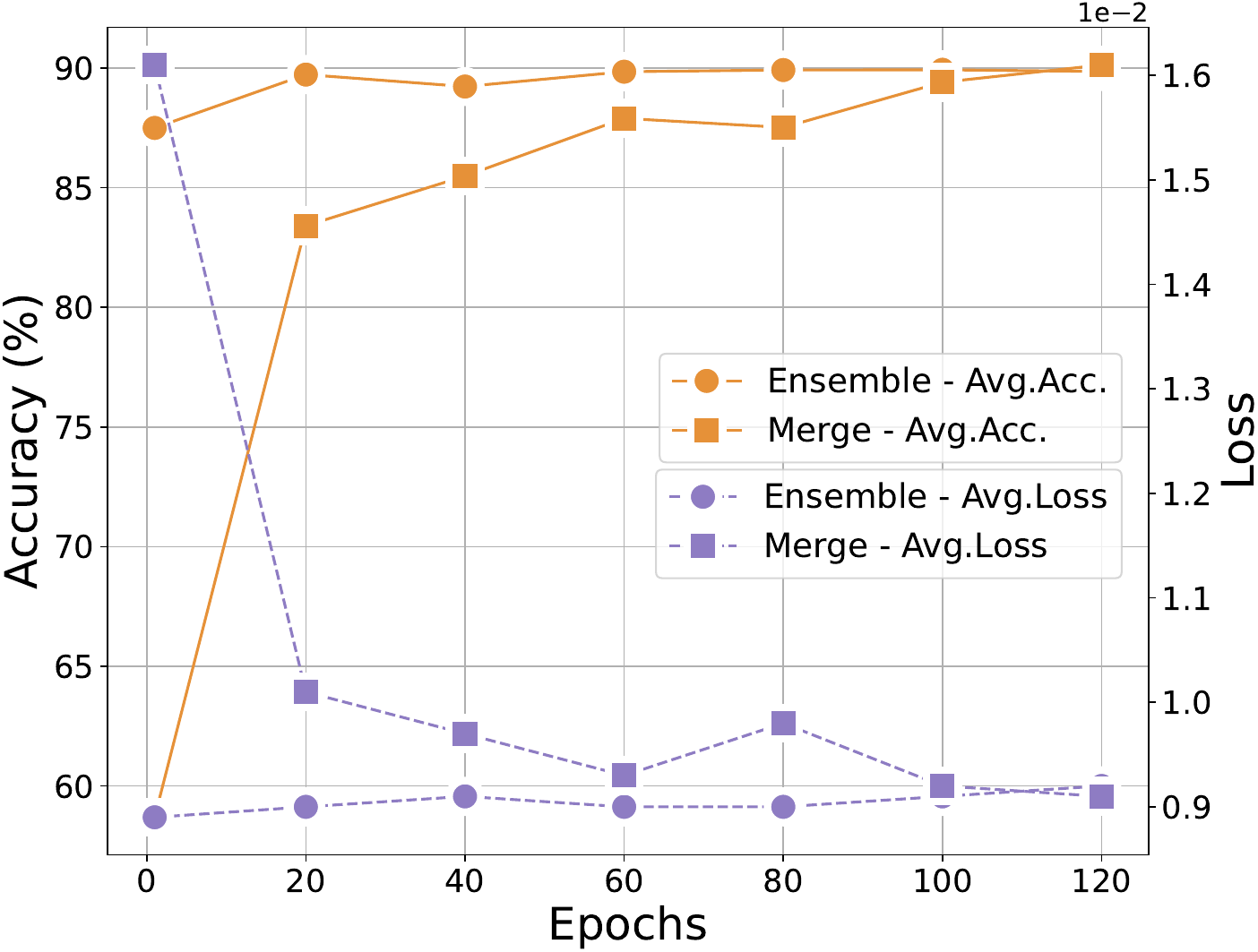}
    \caption{CLIP-RN50 (3 models).}
    \label{fig:short-3}
  \end{subfigure}
  \hfill
  \begin{subfigure}{0.23\linewidth}
    \centering
    \includegraphics[width=\linewidth]{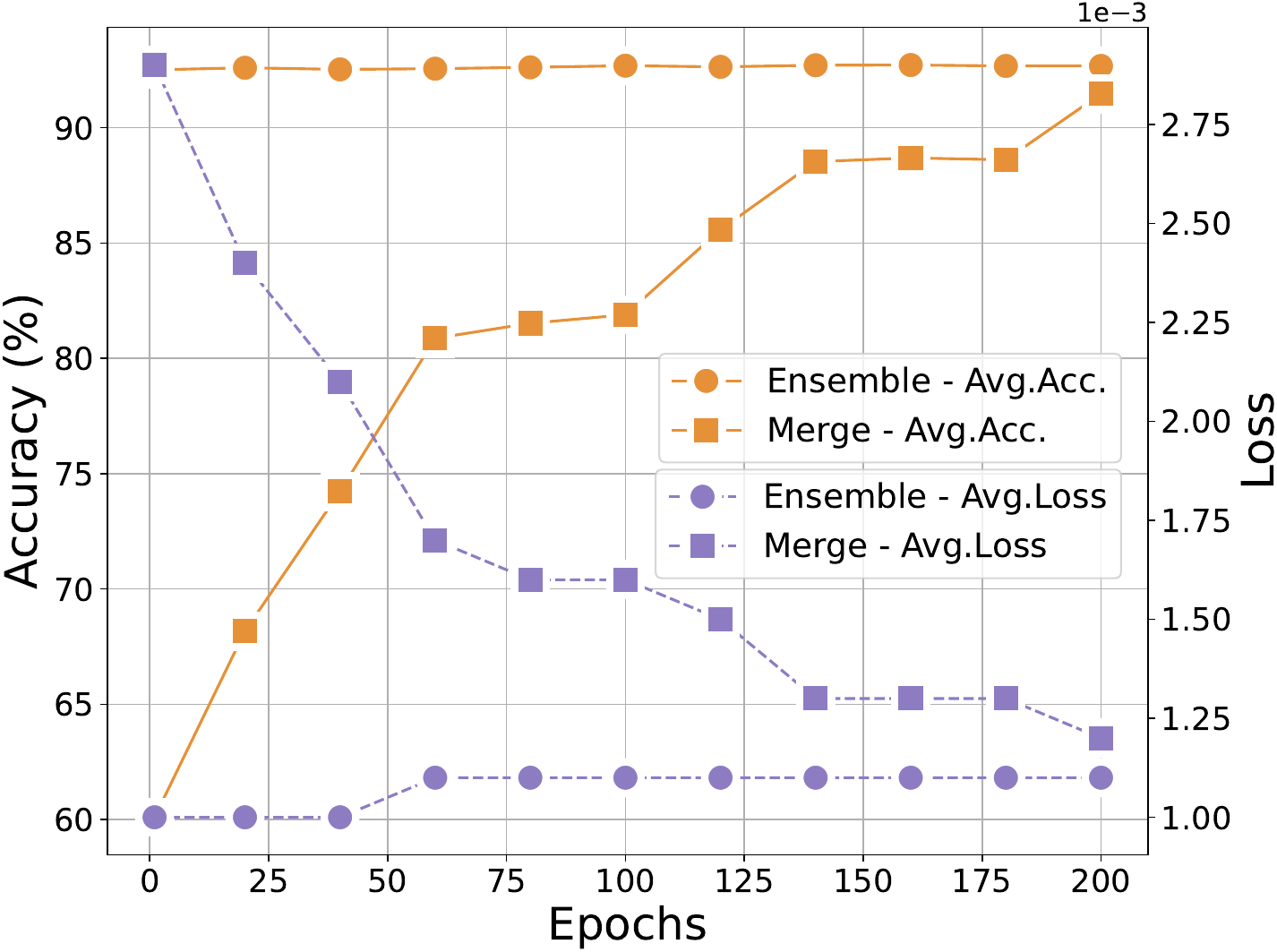}
    \caption{CLIP-RN50 (5 models).}
    \label{fig:short-5}
  \end{subfigure}
  \vspace{-3mm}
  \caption{Training trajectories and final performance for collaborations conducted with varying numbers of models.} 
  \label{fig:trajectory}
  \vspace{-4mm}
\end{figure*}
\noindent\textbf{Training Trajectory of NeuLig. } 
In this experiment, we find that with ViT-based models, training on Portland converges rapidly (within approximately one to two epochs), whereas ResNet-based models converge more slowly. In Figures \ref{fig:short-2}-\ref{fig:short-5}, we illustrate the accuracy and loss trajectories during training when collaborating with 2, 3, and 5 ResNet-based models, respectively, and present the final performance in Figure \ref{fig:trajec}. When collaborating with 2 models, we use the GTSRB and CIFAR100 fine-tuned models; with 3 models, we use the GTSRB, CIFAR100, and RESISC45 fine-tuned models; and with 5 models, we use the same datasets as in Table \ref{tab:datalevel_main}. We can observe that in the early training stage, the performance of merging is significantly lower than that of ensembling. However, as training continues, this gap gradually narrows. The merging performance improves rapidly at first, then slows, and eventually aligns with the ensembling performance.
\begin{figure}[ht]
  \centering
  \begin{subfigure}{0.5\linewidth}
    \centering
    \includegraphics[width=\linewidth]{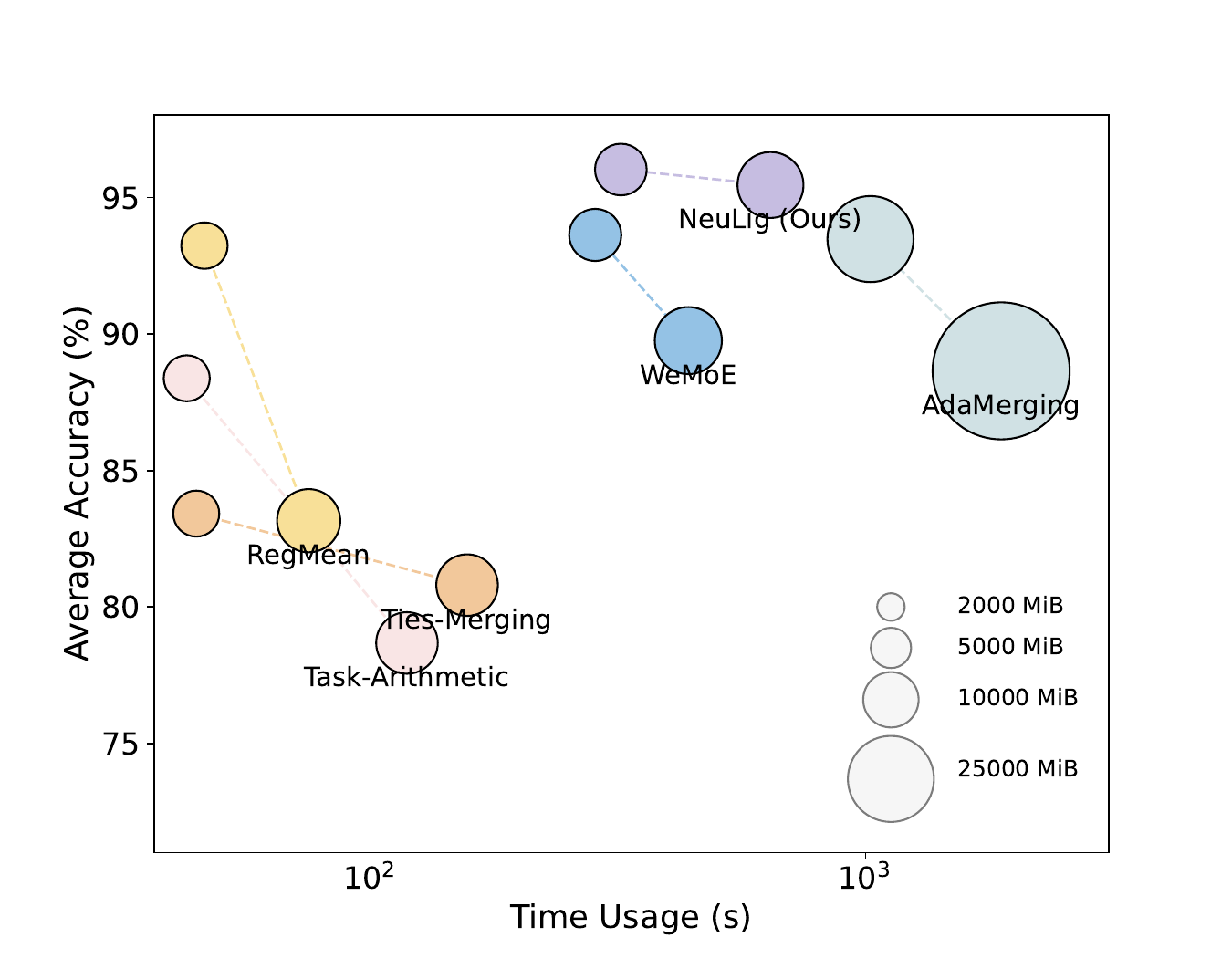}
    \caption{An overall efficiency comparison between \texttt{NeuLig} and other baselines. The size the each colored point represents the VRAM (GPU Memory Usage). Results are conducted under CLIP-ViT-B/32. (Average Accuracy($\uparrow)$, Time Usage($\downarrow$), VRAM($\downarrow$)).}
    \label{fig:efficiency}
  \end{subfigure}
  \hspace{0.02\linewidth} 
  \begin{subfigure}{0.46\linewidth}
    \centering
    \renewcommand{\arraystretch}{0.9} 
    \small
    \caption{Performance when using diverse-origin models.}
    \begin{tabular}{c|c|c}
      \hline
      \multicolumn{3}{c}{\textbf{\# Models = 2}} \\ \hline
      Seperate & \multicolumn{2}{c}{75.22}  \\ \hline
      Random & \multicolumn{2}{c}{6.16} \\ \hline
      \multirow{2}{*}{\texttt{NeuLig}} & \textcolor{teal}{\ding{108}}Ens. & \textcolor{pink}{\ding{117}}Mer. \\ \cline{2-3}
      & 82.19 & 82.17   \\ \hline
      \multicolumn{3}{c}{\textbf{\# Models = 3}} \\ \hline
      Seperate & \multicolumn{2}{c}{65.86}  \\ \hline
      Random & \multicolumn{2}{c}{4.85} \\ \hline
      \multirow{2}{*}{\texttt{NeuLig}} & \textcolor{teal}{\ding{108}}Ens. & \textcolor{pink}{\ding{117}}Mer. \\ \cline{2-3}
      & 70.76 & 69.33   \\ \hline
      \multicolumn{3}{c}{\textbf{\# Models = 5}} \\ \hline
      Seperate & \multicolumn{2}{c}{54.31}  \\ \hline
      Random & \multicolumn{2}{c}{5.11} \\ \hline
      \multirow{2}{*}{\texttt{NeuLig}} & \textcolor{teal}{\ding{108}}Ens. & \textcolor{pink}{\ding{117}}Mer.  \\ \cline{2-3}
      & 52.49 & 50.20  \\ \hline
    \end{tabular}
    \label{fig:diverse-origin}
  \end{subfigure}
  \caption{(a). The efficiency perspective of \texttt{NeuLig}. (b). The performance of \texttt{NeuLig} in the diverse-origin model scenario (2 models: GTSRB-MNIST, 3 models: GTSRB-RESISC45-MNIST).}
  \label{fig:combined}
\end{figure}

\noindent\textbf{Resource Usage of NeuLig. } Figure \ref{fig:efficiency} shows the computing resource consumption and time usage, as well as the average merging accuracy of \texttt{NeuLig} compared to various baselines. Different colors represent different methods, with each color showing two results: the left shows collaboration between two models, and the right indicates collaboration among five models. It can be observed that as the number of models increases, baselines either experience significant performance degradation or a sharp increase in resource usage, while \texttt{NeuLig} shows better resilience.

\noindent\textbf{NeuLig under Diverse-Origin Models. } In this scenario, the models are not fine-tuned from the same pre-trained checkpoint but are instead randomly initialized and trained from scratch. Existing methods fail to function under this condition. However, results when using \texttt{NeuLig} indicates that achieving a considerable performance under this scenario is possible. We present the performance of separate models, random guessing and \texttt{NeuLig} under this scenario in Table \ref{fig:diverse-origin}. Previous works on model merging can achieve performance merely equal to random guessing. In contrast, \texttt{NeuLig} successfully help to maintain performance consistency between merging and ensembling.

\begin{table}[h]
    \centering
    \resizebox{\linewidth}{!}{
        \begin{tabular}{c|c|c|c|c|c|c}
            \hline
            \textbf{Data Scale} & \textbf{Type} & \textbf{0.01} & \textbf{0.05} & \textbf{0.1} & \textbf{0.15} & \textbf{0.2} \\ 
            \hline
            \multirow{2}{*}{CLIP-RN50} & \textcolor{pink}{\ding{117}}Mer.& 80.93 & 88.79 & 90.41 & 90.55 & 91.37 \\ \cline{2-7} 
            & \textcolor{teal}{\ding{108}}Ens. & 90.77 & 91.30 & 91.78 & 91.09 & 90.08 \\
            \hline
            \multirow{2}{*}{CLIP-ViT-B/32} & \textcolor{pink}{\ding{117}}Mer.& 91.43 & 92.96  & 93.46 & 93.84 & 94.26  \\ \cline{2-7} 
            & \textcolor{teal}{\ding{108}}Ens. & 92.60 & 94.24 & 94.47 & 94.31 & 94.62 \\
            \hline
            \textbf{Data Scale} & \textbf{Type} &  \textbf{0.3} & \textbf{0.4} & \textbf{0.6} & \textbf{0.8} & \textbf{1.0} \\ 
            \hline
            \multirow{2}{*}{CLIP-RN50} & \textcolor{pink}{\ding{117}}Mer.& 91.38 & 91.77 & 91.58 & 92.52 & 92.07  \\ \cline{2-7} 
            & \textcolor{teal}{\ding{108}}Ens. & 91.96 & 91.50 & 92.01 & 91.77 & 92.68   \\
            \hline
            \multirow{2}{*}{CLIP-ViT-B/32} & \textcolor{pink}{\ding{117}}Mer.& 94.31 & 94.32 & 94.53 & 94.81 & 95.44  \\ \cline{2-7} 
            & \textcolor{teal}{\ding{108}}Ens. & 94.87 & 94.94 & 94.96 & 95.08 & 95.46  \\
            \hline
        \end{tabular}
    }
    \caption{The performance variation of \texttt{NeuLig} under the semi-supervised learning setup when datasets of different scales are used. For instance, a data scale of 0.3 indicates that 30\% of the unlabeled test dataset is used.}
    \label{tab:data_scale}
\end{table}

\noindent\textbf{NeuLig with Varying Dataset Scales. } In Table \ref{tab:data_scale}, we explore the impact of the available data scale. We employ ten different data scales from 0.01 to 1.0 to assess the impact. It can be observed that as the available data scale decreases, the performance of merging and ensembling remains largely stable until the data scale becomes extremely small (e.g., 0.05), and the performance consistency between merging and ensembling persists at different data scales.
\section{Conclusion}
\label{sec:conclusion}
\begingroup
\xspaceskip=3pt    
In this work, we explore an interesting possibility: whether performance consistency between parameter-level merging and prediction-level ensembling can be achieved in the multi-model collaboration scenario. Through theoretical analysis, we provide an affirmative answer to this question and propose a validation framework named \texttt{NeuLig} to verify the practical feasibility of our findings. We also conduct in-depth discussions on various properties of the achieved performance consistency using CoopVec Map and provide detailed analysis. Experimental results on ResNet-based models and ViT-based models demonstrate that \texttt{NeuLig} effectively helps to validate the performance consistency between parameter-level merging and prediction-level ensembling regardless of model scale and quantity, providing new insights into multi-model collaboration.
\par
\endgroup

\setcounter{page}{0}
\setcounter{section}{0}

\onecolumn
\title{}
{
    \centering
    \Large
    \textbf{\thetitle}\\
}
\begin{center}
\Large
\textit{- Supplementary Material -}
\end{center}

\section{NeuLig under More Models}
In Table \ref{tab:datalevel_main}, we evaluate the performance of \texttt{NeuLig} in scenarios involving the collaboration of up to five models. To further explore its scalability, we extend this investigation to scenarios with a greater number of models. Specifically, we incorporate two additional models fine-tuned on the STL10 and SVHN datasets, increasing the total to seven models. The experimental results are presented in Table \ref{tab:datalevel_main_7models}.

\begin{table*}[ht!]
    \centering
    \scriptsize
    \renewcommand{\arraystretch}{1.3}
    \setlength{\tabcolsep}{3pt}
    \begin{subtable}[t]{\textwidth}
        \centering
        \resizebox{\textwidth}{!}{
            \begin{tabular}{lcccccccccccccccccccccccccccccccc}
                \toprule
                \textbf{Method} & 
                \multicolumn{3}{c}{\textbf{GTSRB}} & 
                \multicolumn{3}{c}{\textbf{CIFAR100}} & 
                \multicolumn{3}{c}{\textbf{RESISC45}} & 
                \multicolumn{3}{c}{\textbf{CIFAR10}} & 
                \multicolumn{3}{c}{\textbf{MNIST}} & 
                \multicolumn{3}{c}{\textbf{STL10}} & 
                \multicolumn{3}{c}{\textbf{SVHN}} &
                \multicolumn{3}{c}{\textbf{Avg}} \\
                \midrule
                Pre-trained & \multicolumn{3}{c}{32.56} & \multicolumn{3}{c}{64.20} & \multicolumn{3}{c}{60.22} & \multicolumn{3}{c}{89.83} & \multicolumn{3}{c}{48.25} & \multicolumn{3}{c}{15.91} & \multicolumn{3}{c}{8.31} & \multicolumn{3}{c}{45.61} \\
                Fine-tuned & \multicolumn{3}{c}{98.95} & \multicolumn{3}{c}{84.22} & \multicolumn{3}{c}{94.13} & \multicolumn{3}{c}{97.13} & \multicolumn{3}{c}{99.56} & \multicolumn{3}{c}{96.09} & \multicolumn{3}{c}{96.80} & \multicolumn{3}{c}{95.27} \\
                \midrule
                & \textbf{Mer. $\uparrow$} & \textbf{Ens. $\uparrow$} & \textbf{Gap $ \downarrow $} &
                \textbf{Mer. $\uparrow$} & \textbf{Ens. $\uparrow$} & \textbf{Gap $ \downarrow $} & 
                \textbf{Mer. $\uparrow$} & \textbf{Ens. $\uparrow$} & \textbf{Gap $ \downarrow $} & 
                \textbf{Mer. $\uparrow$} & \textbf{Ens. $\uparrow$} & \textbf{Gap $ \downarrow $} & 
                \textbf{Mer. $\uparrow$} & \textbf{Ens. $\uparrow$} & \textbf{Gap $ \downarrow $} & 
                \textbf{Mer. $\uparrow$} & \textbf{Ens. $\uparrow$} & \textbf{Gap $ \downarrow $} &
                \textbf{Mer. $\uparrow$} & \textbf{Ens. $\uparrow$} & \textbf{Gap $ \downarrow $} &
                \textbf{Mer. $\uparrow$} & \textbf{Ens. $\uparrow$} & \textbf{Gap $ \downarrow $} \\
                \cmidrule(lr){2-4} \cmidrule(lr){5-7} 
                \cmidrule(lr){8-10} \cmidrule(lr){11-13} \cmidrule(lr){14-16} \cmidrule(lr){17-19} \cmidrule(lr){20-22} \cmidrule(lr){23-25}
                &&&&&&&&&&\multicolumn{8}{l}{\textit{Multi-Task Model Collaboration Methods}}&& \\
                Simple-Averaging\cite{wortsman2022model} & 53.78 & 85.95 & 32.17 & 73.27 & 74.92 & 1.65 & 68.97 & 81.14 & 12.17 & 94.40 & 96.78 & 2.38 & 83.04 & 97.65 & 14.61 & 41.65 & 88.06 & 46.41 & 14.99 & 95.99 & 81.00 & 61.44 & 88.64 & 27.20 \\
                Task-Arithmetic\cite{ilharcoediting} & 57.97 & 88.30 & 30.33 & 62.29 & 75.92 & 13.63 & 53.60 & 73.90 & 20.30 & 91.84 & 96.56 & 4.72 & 89.99 & 99.22 & 9.23 & 67.64 & 90.58 & 22.94 & 29.84 & 94.07 & 64.23 & 64.74 & 88.36 & 23.62 \\
                Ties-Merging\cite{yadav2024ties} & 65.17 & - & - & 71.14 & - & - & 69.33 & - & - & 94.63 & - & - & 91.73 & - & - & 61.32 & - & - & 22.81 & - & - & 68.02 & - & -\\
                RegMean\cite{jin2023dataless} & 64.17 & - & - & 73.12 & - & - & 76.22 & - & - & 94.81 & - & - & 89.63 & - & - & 62.80 & - & - & 17.29 & - & - &  68.29 & - & - \\
                AdaMerging\cite{yangadamerging} & 90.92 & 92.34 & 1.42 & 69.92 & 76.00 & 6.08 & 84.51 & 83.30 & 1.21 & 92.65 & 96.58 & 3.93 & 97.25 & 98.38 & 1.13 & 96.67 & 90.65 & 6.02 & 10.85 & 96.45 & 85.60 & 77.54 & 90.53 & 12.99 \\
                WeMoE\cite{tangmerging} & 91.36 & 92.80 & 1.44 & 72.45 & 74.30 & 1.85 & 86.50 & 86.98 & 0.48 & 94.24 & 96.58 & 2.34 & 97.80 & 98.12 & 0.32 & 93.48 & 97.52 & 4.04 & 26.48 & 96.53 & 70.05 & 80.33 & 91.83 & 11.50 \\
                \midrule
                &&&&&&&&&&&\multicolumn{3}{l}{\textit{Neural Ligand}}&&&&&& \\
                \cellcolor{teal!10}Ours (Semi-Supervised) & \cellcolor{teal!10}99.05 & \cellcolor{teal!10}99.10 & \cellcolor{orange!10}\textbf{0.05} & \cellcolor{teal!10}85.39 & \cellcolor{teal!10}85.62 & \cellcolor{orange!10}0.23 & \cellcolor{teal!10}93.87 & \cellcolor{teal!10}94.05 & \cellcolor{orange!10}\textbf{0.18} & \cellcolor{teal!10}\textbf{96.33} & \cellcolor{teal!10}\textbf{96.78} & \cellcolor{orange!10}0.45 & \cellcolor{teal!10}\textbf{99.58} & \cellcolor{teal!10}99.57 & \cellcolor{orange!10}\textbf{0.01} & \cellcolor{teal!10}\textbf{96.94} & \cellcolor{teal!10}96.08 & \cellcolor{orange!10}0.86 & \cellcolor{teal!10}96.82 & \cellcolor{teal!10}96.79 & \cellcolor{orange!10}\textbf{0.03} & \cellcolor{teal!10}95.43 \textcolor{teal}{(+15.10)} & \cellcolor{teal!10}95.43 \textcolor{teal}{(+3.60)} & \cellcolor{orange!10}\textbf{0.00} \textcolor{orange}{(-11.50)} \\
                \cellcolor{teal!10}Ours (Supervised) & \cellcolor{teal!10}\textbf{99.20} & \cellcolor{teal!10}\textbf{99.33} & \cellcolor{orange!10}0.13 & \cellcolor{teal!10}\textbf{87.26} & \cellcolor{teal!10}\textbf{87.44} & \cellcolor{orange!10}\textbf{0.18} & \cellcolor{teal!10}\textbf{94.02} & \cellcolor{teal!10}\textbf{94.38} & \cellcolor{orange!10}0.36 & \cellcolor{teal!10}96.10 & \cellcolor{teal!10}96.32 & \cellcolor{orange!10}\textbf{0.22} & \cellcolor{teal!10}99.44 & \cellcolor{teal!10}\textbf{99.87} & \cellcolor{orange!10}0.43 & \cellcolor{teal!10}96.35 & \cellcolor{teal!10}\textbf{96.48} & \cellcolor{orange!10}\textbf{0.13} & \cellcolor{teal!10}\textbf{96.88} & \cellcolor{teal!10}\textbf{97.20} & \cellcolor{orange!10}0.32 & \cellcolor{teal!10}\textbf{95.61} \textcolor{teal}{(+15.28)} & \cellcolor{teal!10}\textbf{95.86} \textcolor{teal}{(+4.03)} & \cellcolor{orange!10}0.09 \textcolor{orange}{(-11.41)}\\
                \bottomrule
            \end{tabular}
        }
    \end{subtable}
    \caption{Results of various methods across multiple datasets, including the merging performance, the ensembling performance, and the performance gap for CLIP-ViT-B/32.}
    \label{tab:datalevel_main_7models}
\end{table*}
As observed, even with an increased number of collaborating models, \texttt{NeuLig} consistently demonstrates exceptionally low performance gaps while significantly outperforming baseline methods. Remarkably, under the semi-supervised setting, the performance gap is entirely eliminated. These results further reinforce the validity of our findings and affirm the effectiveness of \texttt{NeuLig} as a robust validation framework.

\label{sec:rationale}

\section{NeuLig under Other Model Types}
\begin{table*}[ht]
    \centering
    \scriptsize
    \renewcommand{\arraystretch}{1.3}
    \setlength{\tabcolsep}{3pt}
    \begin{subtable}[t]{\textwidth}
        \centering
        \resizebox{\textwidth}{!}{
            \begin{tabular}{lcccccccccccccccccc}
                \toprule
                \textbf{Method} & 
                \multicolumn{3}{c}{\textbf{GTSRB}} & 
                \multicolumn{3}{c}{\textbf{CIFAR100}} & 
                \multicolumn{3}{c}{\textbf{RESISC45}} & 
                \multicolumn{3}{c}{\textbf{CIFAR10}} & 
                \multicolumn{3}{c}{\textbf{MNIST}} & 
                \multicolumn{3}{c}{\textbf{Avg}} \\
                \midrule
                Pre-trained & \multicolumn{3}{c}{50.55} & \multicolumn{3}{c}{75.82} & \multicolumn{3}{c}{71.33} & \multicolumn{3}{c}{95.57} & \multicolumn{3}{c}{76.36} & \multicolumn{3}{c}{73.93} \\
                Fine-tuned & \multicolumn{3}{c}{99.11} & \multicolumn{3}{c}{91.64} & \multicolumn{3}{c}{96.05} & \multicolumn{3}{c}{98.80} & \multicolumn{3}{c}{99.70} & \multicolumn{3}{c}{97.06} \\
                \midrule
                & \textbf{Mer. $\uparrow$} & \textbf{Ens. $\uparrow$} & \textbf{Gap $ \downarrow $} &
                \textbf{Mer. $\uparrow$} & \textbf{Ens. $\uparrow$} & \textbf{Gap $ \downarrow $} & 
                \textbf{Mer. $\uparrow$} & \textbf{Ens. $\uparrow$} & \textbf{Gap $ \downarrow $} & 
                \textbf{Mer. $\uparrow$} & \textbf{Ens. $\uparrow$} & \textbf{Gap $ \downarrow $} & 
                \textbf{Mer. $\uparrow$} & \textbf{Ens. $\uparrow$} & \textbf{Gap $ \downarrow $} & 
                \textbf{Mer. $\uparrow$} & \textbf{Ens. $\uparrow$} & \textbf{Gap $ \downarrow $}  \\
                \cmidrule(lr){2-4} \cmidrule(lr){5-7} 
                \cmidrule(lr){8-10} \cmidrule(lr){11-13} \cmidrule(lr){14-16} \cmidrule(lr){17-19}
                &&&&&&&\multicolumn{6}{l}{\textit{Multi-Task Model Collaboration Methods}}&&&& \\
                Simple-Averaging\cite{wortsman2022model} & 67.48 & 94.23 & 26.75 & 86.26 & 88.89 & 3.37 & 80.76 & 90.42 & 9.66 & 94.26 & 96.48 & 2.22 & 93.26 & 98.84 & 5.58 & 84.40 & 93.77 & 9.37 \\
                Task-Arithmetic\cite{ilharcoediting} & 68.23 & 94.15 & 25.92 & 85.46 & 89.13 & 3.67 & 80.48 & 90.91 & 10.43 & 93.92 & 97.56 & 3.64 & 93.78 & 98.92 & 5.14 & 84.37 & 94.13 & 9.76\\
                Ties-Merging\cite{yadav2024ties} & 71.68 & - & - & 85.64 & - & - & 86.74 & - & - & 95.39 & - & - & 91.93 & - & - & 86.28 & - & -\\
                RegMean\cite{jin2023dataless} & 84.57 & - & - & 87.72 & - & - & 90.40 & - & - & 98.59 & - & - & 99.02 & - & - & 92.06 & - & - \\
                AdaMerging\cite{yangadamerging} & 97.78 & 98.65 & 0.87 & 83.02 & 84.43 & 1.41 & 92.66 & 97.89 & 5.23 & 97.12 & 98.83 & 1.71 & 94.29 & 97.23 & 2.94 & 93.17 & 95.61 & 2.44 \\
                WeMoE\cite{tangmerging} & 97.90 & 98.56 & 0.66 & 85.86 & 87.22 & 1.36 & 92.69 & 95.43 & 2.74 & 96.97 & 98.71 & 1.74 & 97.44 & 98.80 & 1.36 & 94.17 & 95.74 & 1.57 \\
                \midrule
                &&&&&&&&\multicolumn{3}{l}{\textit{Neural Ligand}}&&&&&& \\
                \cellcolor{teal!10}Ours (Semi-Supervised) & \cellcolor{teal!10}\textbf{99.90} & \cellcolor{teal!10}\textbf{99.92} & \cellcolor{orange!10}\textbf{0.02} & \cellcolor{teal!10}\textbf{91.42} & \cellcolor{teal!10}\textbf{91.36} & \cellcolor{orange!10}\textbf{0.06} & \cellcolor{teal!10}\textbf{96.54} & \cellcolor{teal!10}96.60 & \cellcolor{orange!10}\textbf{0.06} & \cellcolor{teal!10}98.97 & \cellcolor{teal!10}99.12 & \cellcolor{orange!10}0.15 & \cellcolor{teal!10}\textbf{99.88} & \cellcolor{teal!10}\textbf{99.88} & \cellcolor{orange!10}\textbf{0.00} & \cellcolor{teal!10}\textbf{97.34} \textcolor{teal}{(+3.17)} & \cellcolor{teal!10}97.38 \textcolor{teal}{(+1.64)} & \cellcolor{orange!10}\textbf{0.04} \textcolor{orange}{(-1.53)} \\
                \cellcolor{teal!10}Ours (Supervised) & \cellcolor{teal!10}99.86 & \cellcolor{teal!10}99.90 & \cellcolor{orange!10}0.04 & \cellcolor{teal!10}91.02 & \cellcolor{teal!10}91.34 & \cellcolor{orange!10}0.32 & \cellcolor{teal!10}96.42 & \cellcolor{teal!10}\textbf{96.65} & \cellcolor{orange!10}0.23 & \cellcolor{teal!10}\textbf{99.68} & \cellcolor{teal!10}\textbf{99.62} & \cellcolor{orange!10}\textbf{0.06} & \cellcolor{teal!10}99.73 & \cellcolor{teal!10}99.65 & \cellcolor{orange!10}0.08 & \cellcolor{teal!10}\textbf{97.34} \textcolor{teal}{(+3.17)} & \cellcolor{teal!10}\textbf{97.43} \textcolor{teal}{(+1.69)} & \cellcolor{orange!10}0.09 \textcolor{orange}{(-1.48)}\\
                \bottomrule
            \end{tabular}
        }
    \end{subtable}
    \caption{Results of various methods across multiple datasets, including the merging performance, the ensembling performance, and the performance gap for CLIP-ViT-L/14.}
    \label{tab:datalevel_main_l14}
\end{table*}
In the main manuscript, we employ two model architectures: CLIP-RN50 and CLIP-ViT-B/32. To further investigate the effectiveness of \texttt{NeuLig} with larger model architectures, we conduct additional experiments using CLIP-ViT-L/14 as the backbone. The results of these experiments are summarized in Table \ref{tab:datalevel_main_l14}. It is evident that when using larger model architectures, \texttt{NeuLig} remains a highly effective validation framework. All baseline methods continue to exhibit relatively large performance gaps to varying degrees, whereas \texttt{NeuLig} consistently demonstrates minimal performance differences.

\begin{figure*}[ht]
    \centering
    \begin{minipage}{\textwidth}
    \captionsetup[subfigure]{labelformat=empty} 
        \begin{subfigure}{\textwidth}
            \centering
            \begin{subfigure}{0.2\textwidth}
                \centering
                \includegraphics[width=\linewidth]{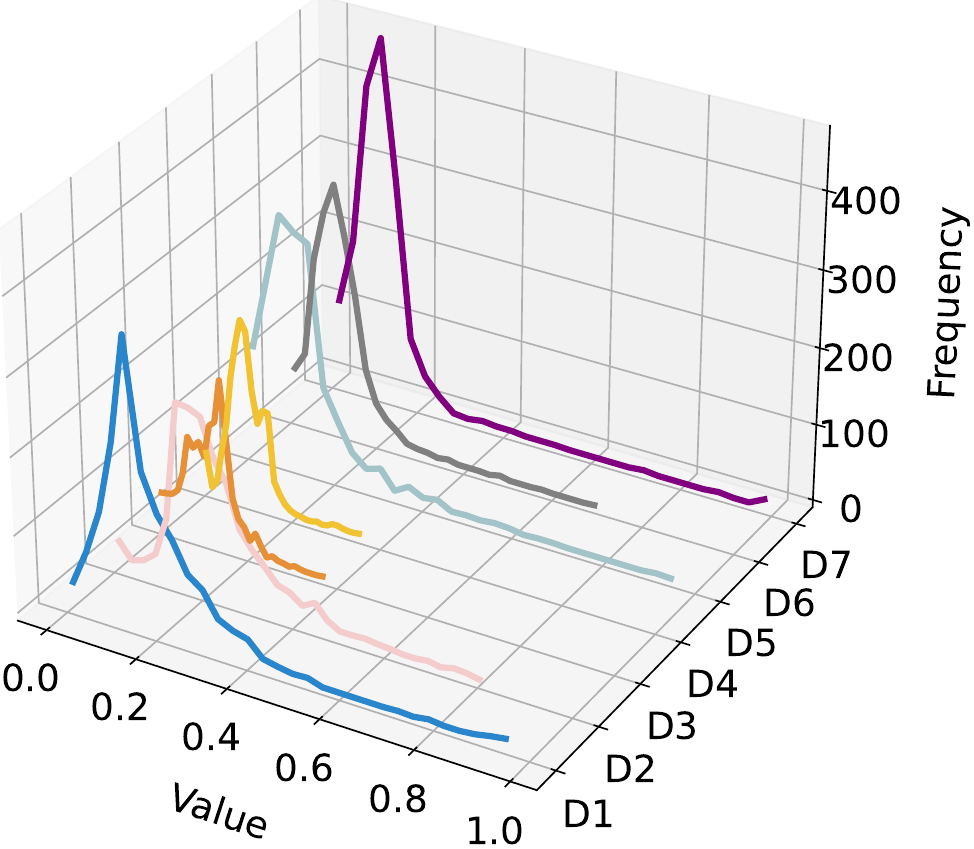} 
            \caption{GTSRB}
            \end{subfigure}
            \begin{subfigure}{0.2\textwidth}
                \centering
                \includegraphics[width=\linewidth]{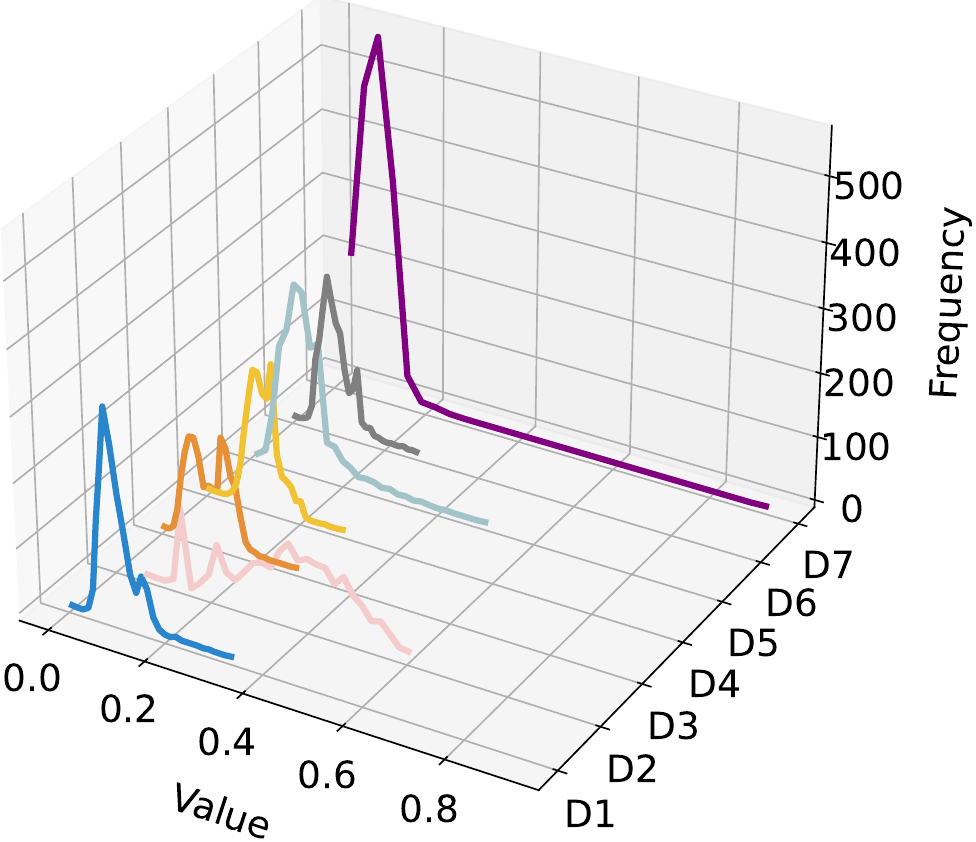} 
            \caption{CIFAR100}
            \end{subfigure}
            \begin{subfigure}{0.2\textwidth}
                \centering
                \includegraphics[width=\linewidth]{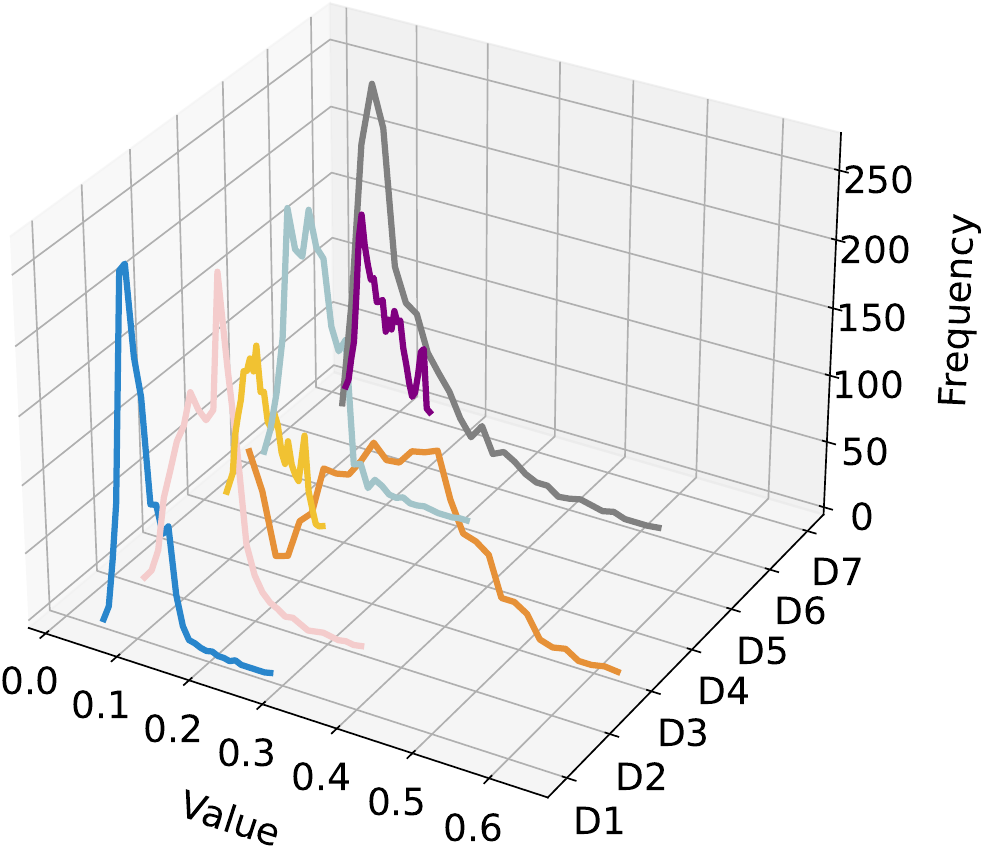} 
                \caption{RESISC45}
            \end{subfigure}
            \begin{subfigure}{0.2\textwidth}
                \centering
                \includegraphics[width=\linewidth]{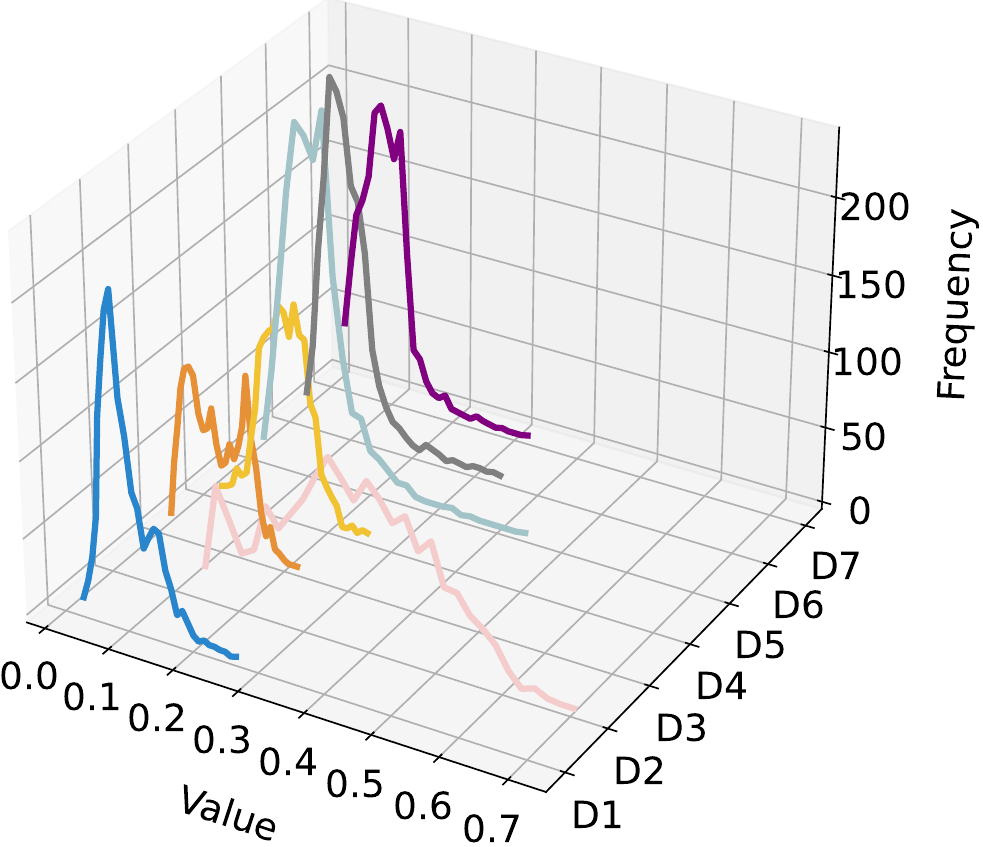}
                \caption{CIFAR10}
            \end{subfigure}
            \begin{subfigure}{0.2\textwidth}
                \centering
                \includegraphics[width=\linewidth]{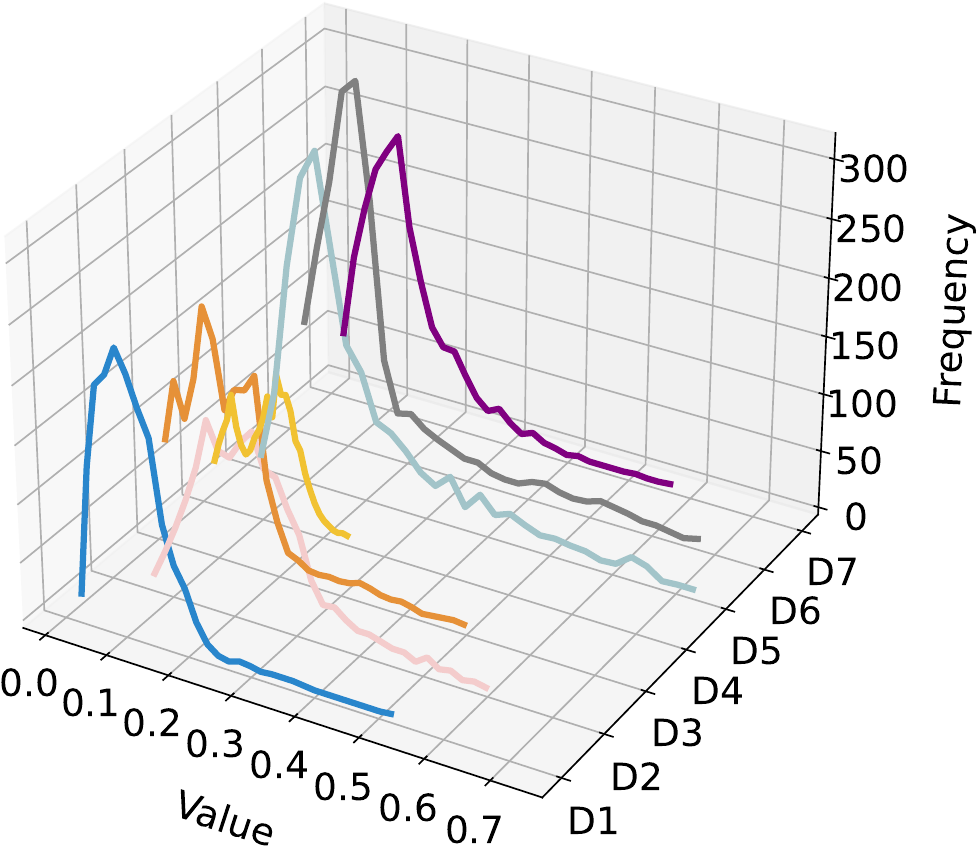}
                \caption{MNIST}
            \end{subfigure}
            \begin{subfigure}{0.2\textwidth}
                \centering
                \includegraphics[width=\linewidth]{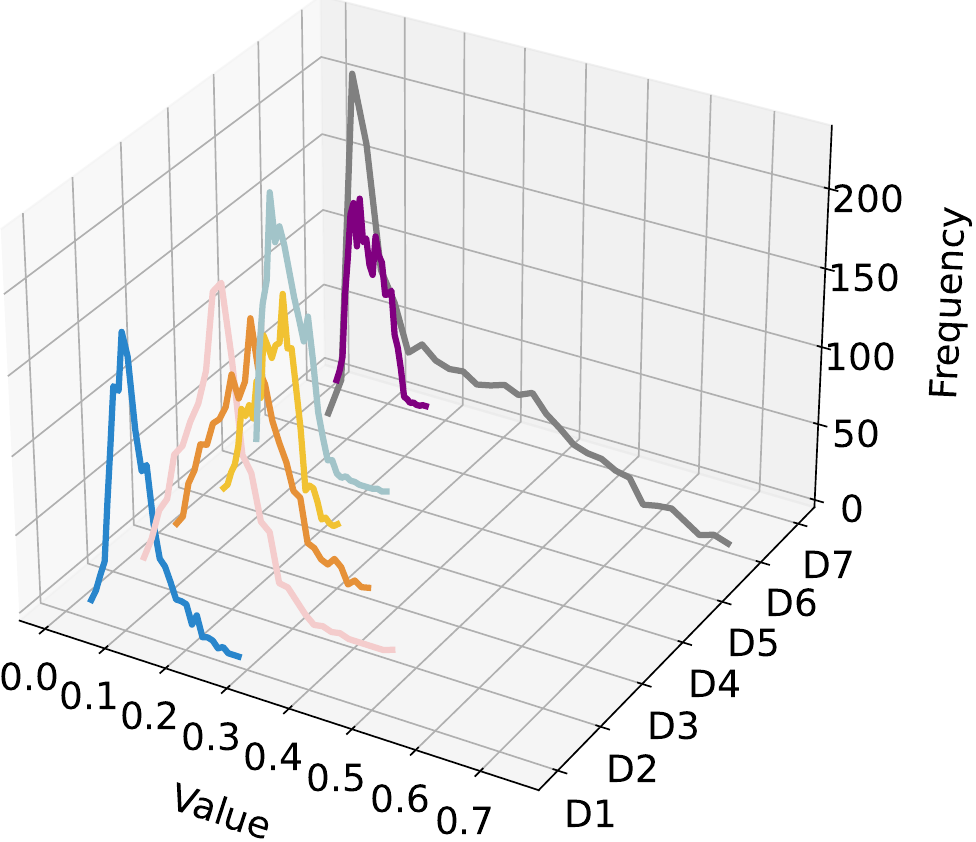} 
                \caption{STL10}
            \end{subfigure}
            \begin{subfigure}{0.2\textwidth}
                \centering
                \includegraphics[width=\linewidth]{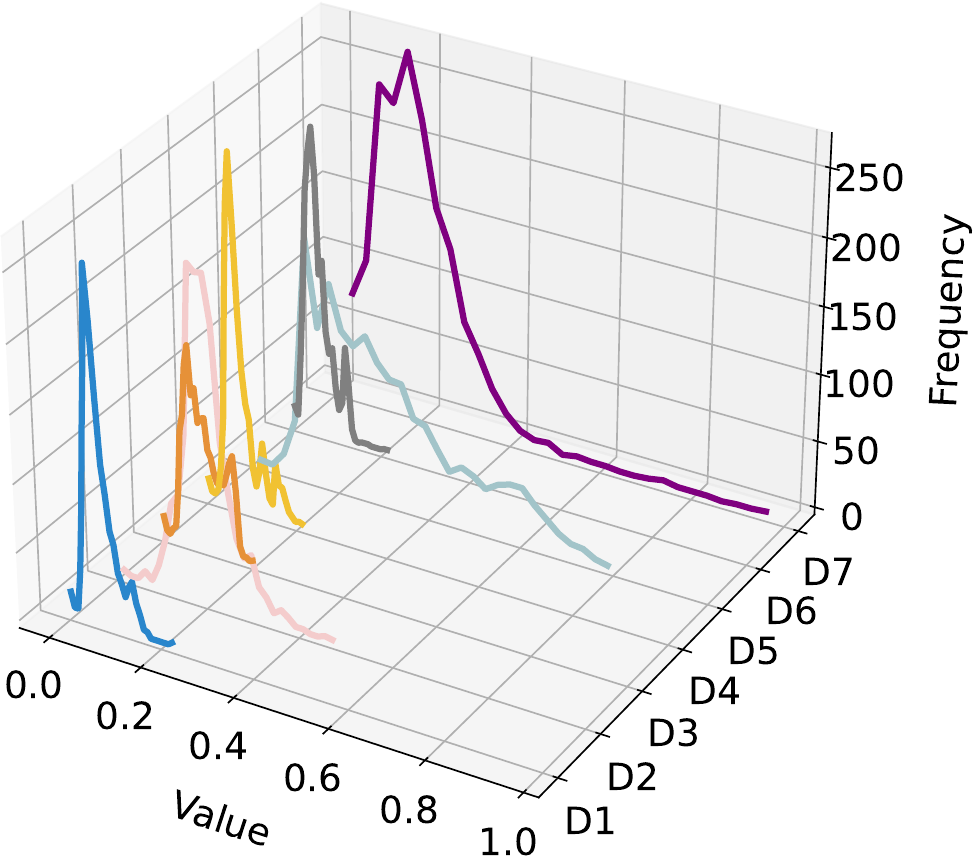}
                \caption{SVHN}
            \end{subfigure}
            \begin{subfigure}{0.2\textwidth}
            \centering
            \includegraphics[width=\linewidth]{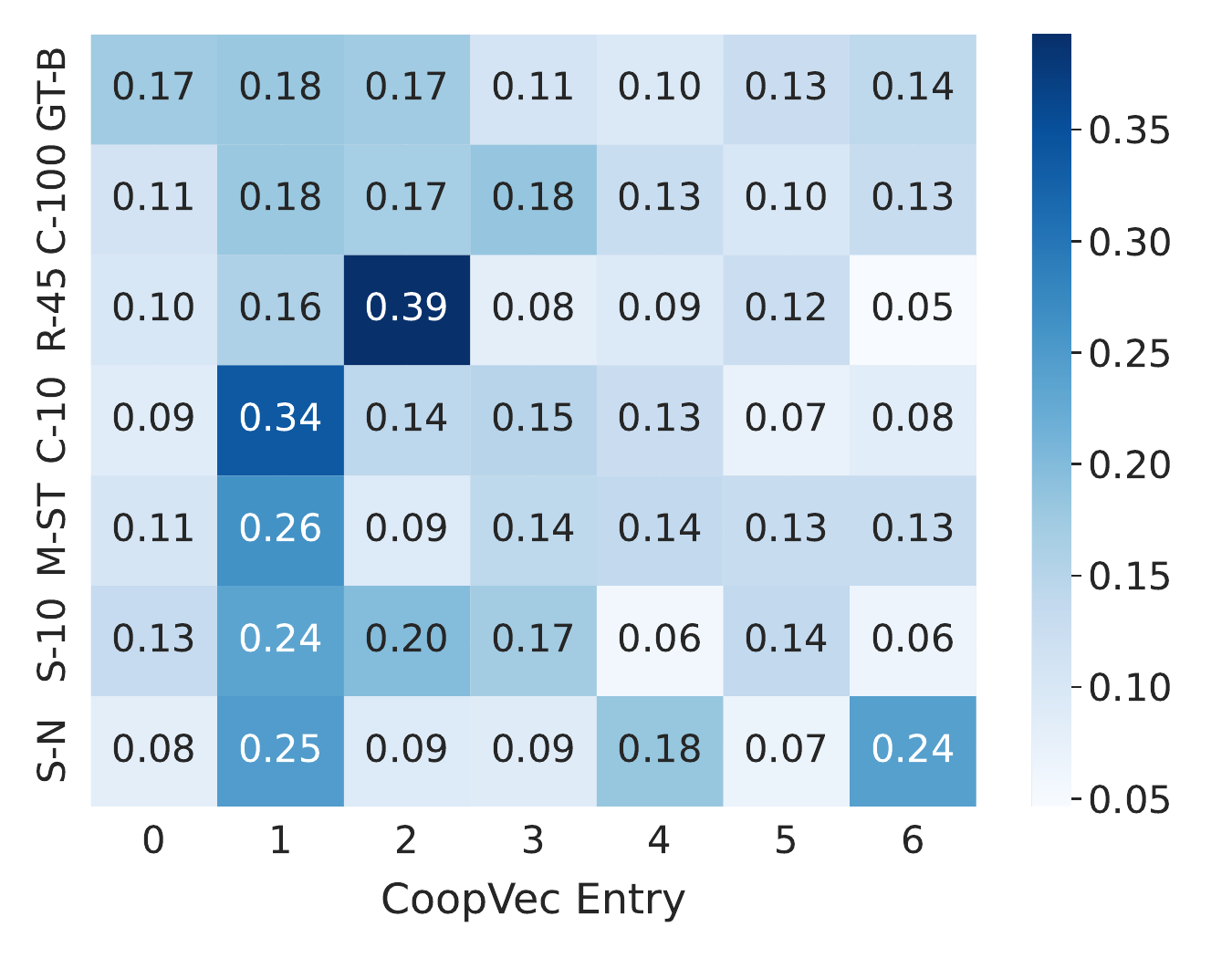}
            \caption{CoopVec Map}
            \end{subfigure}
        \caption{(a) CLIP-RN50.}
        \end{subfigure}        
        
        \begin{subfigure}{\textwidth}
            \centering
            \begin{subfigure}{0.2\textwidth}
                \centering
                \includegraphics[width=\linewidth]{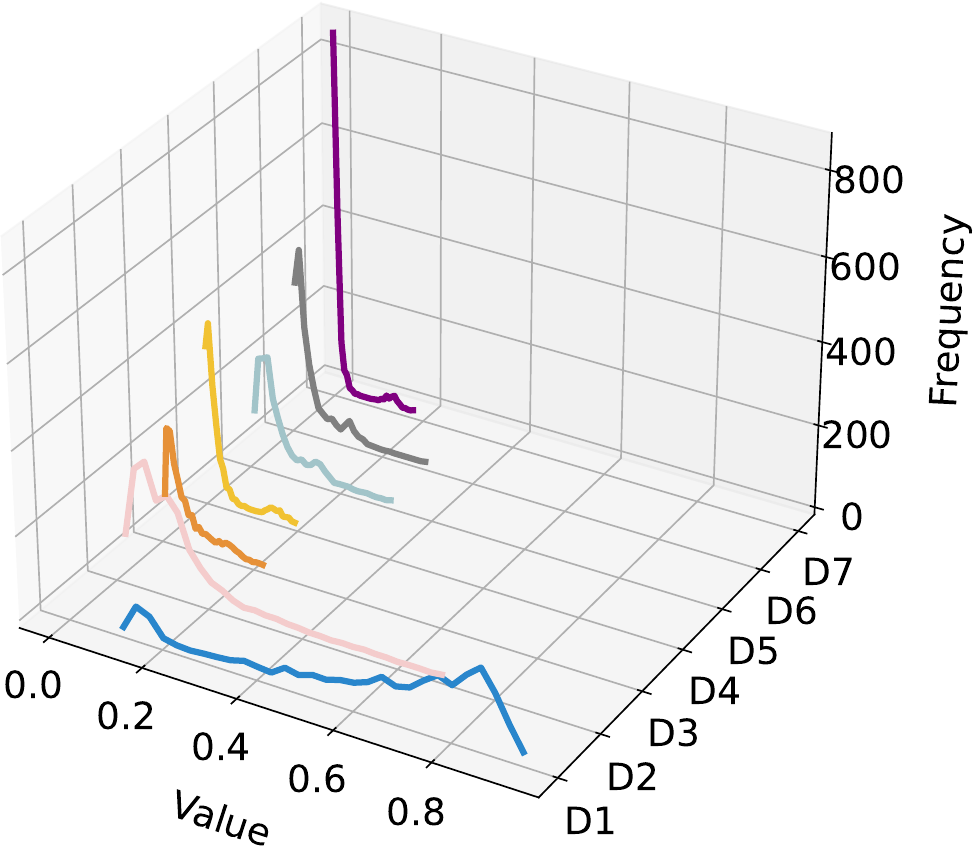} 
                \caption{GTSRB}
            \end{subfigure}
            \begin{subfigure}{0.2\textwidth}
                \centering
                \includegraphics[width=\linewidth]{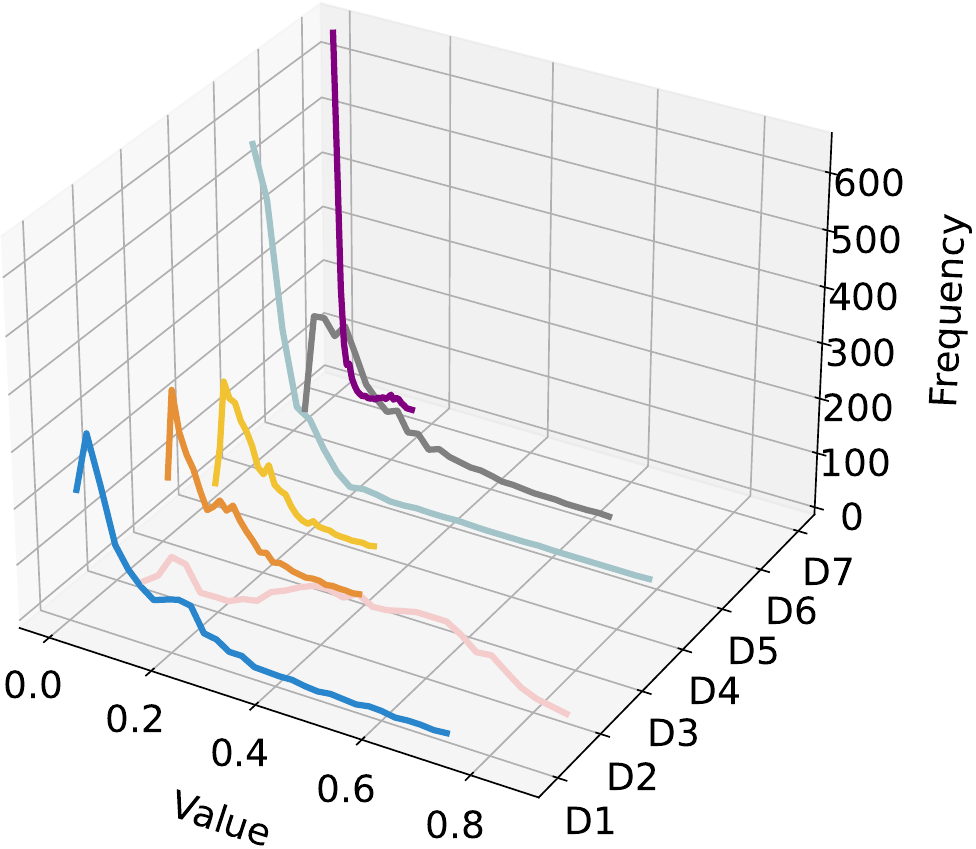} 
                \caption{CIFAR100}
            \end{subfigure}
            \begin{subfigure}{0.2\textwidth}
                \centering
                \includegraphics[width=\linewidth]{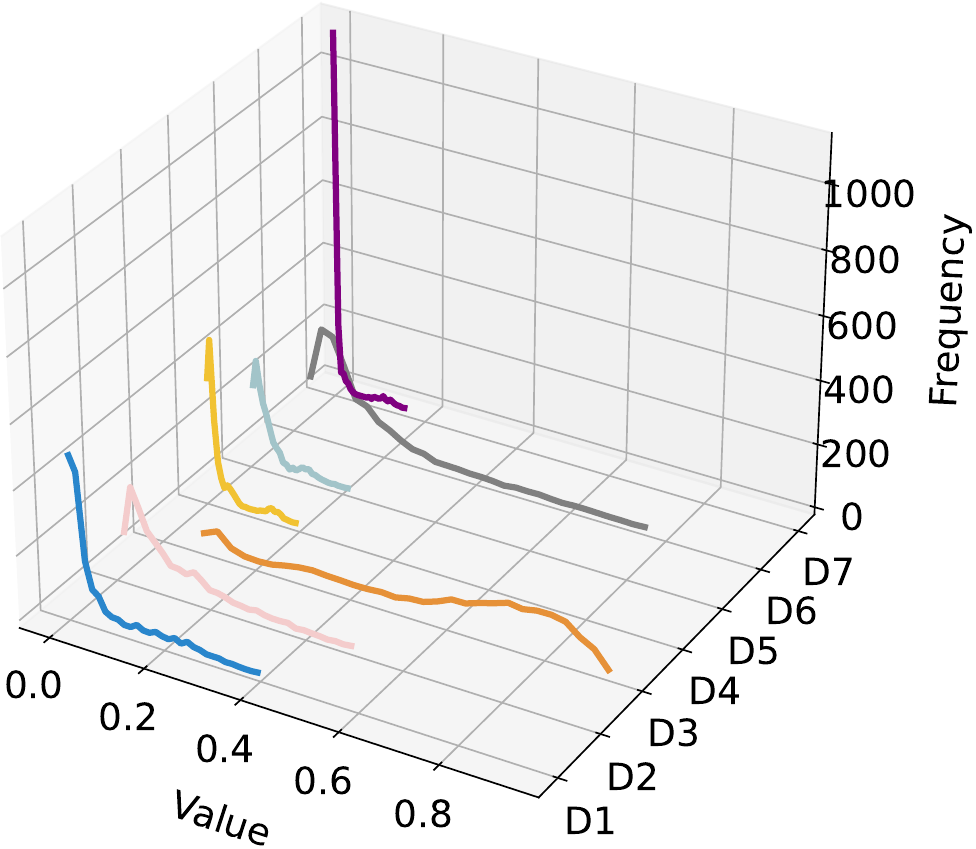} 
                \caption{RESISC45}
            \end{subfigure}
            \begin{subfigure}{0.2\textwidth}
                \centering
                \includegraphics[width=\linewidth]{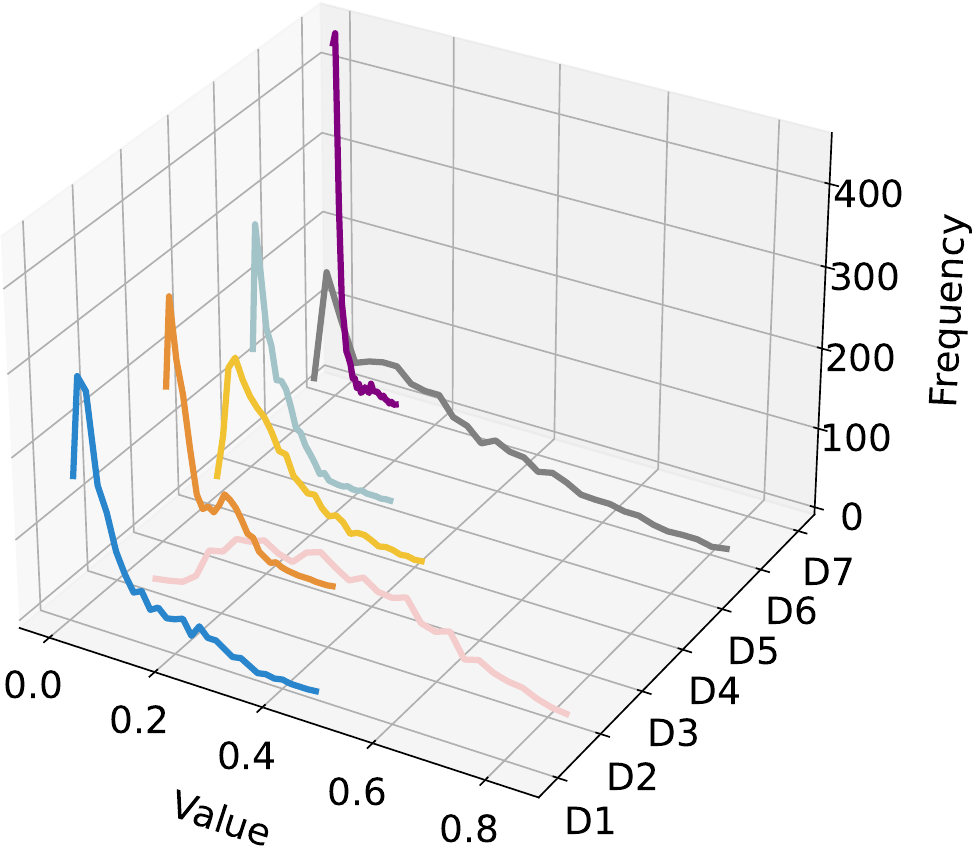} 
                \caption{CIFAR10}
            \end{subfigure}
            \begin{subfigure}{0.2\textwidth}
                \centering
                \includegraphics[width=\linewidth]{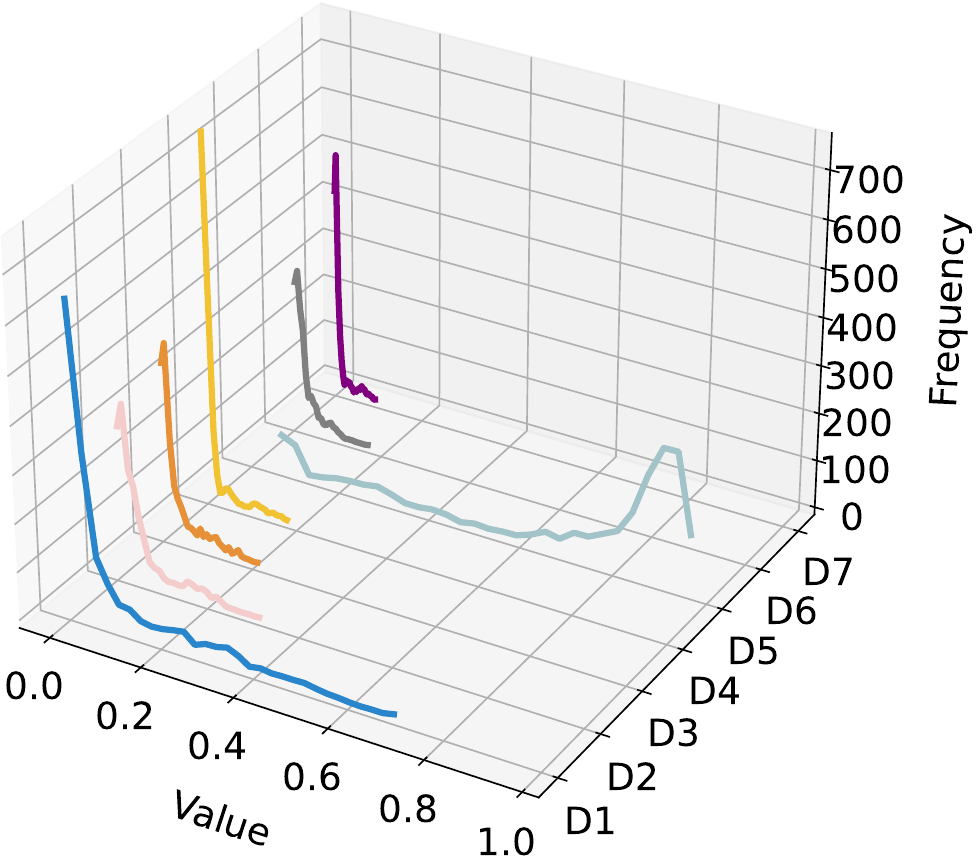} 
                \caption{MNIST}
            \end{subfigure}
            \begin{subfigure}{0.2\textwidth}
                \centering
                \includegraphics[width=\linewidth]{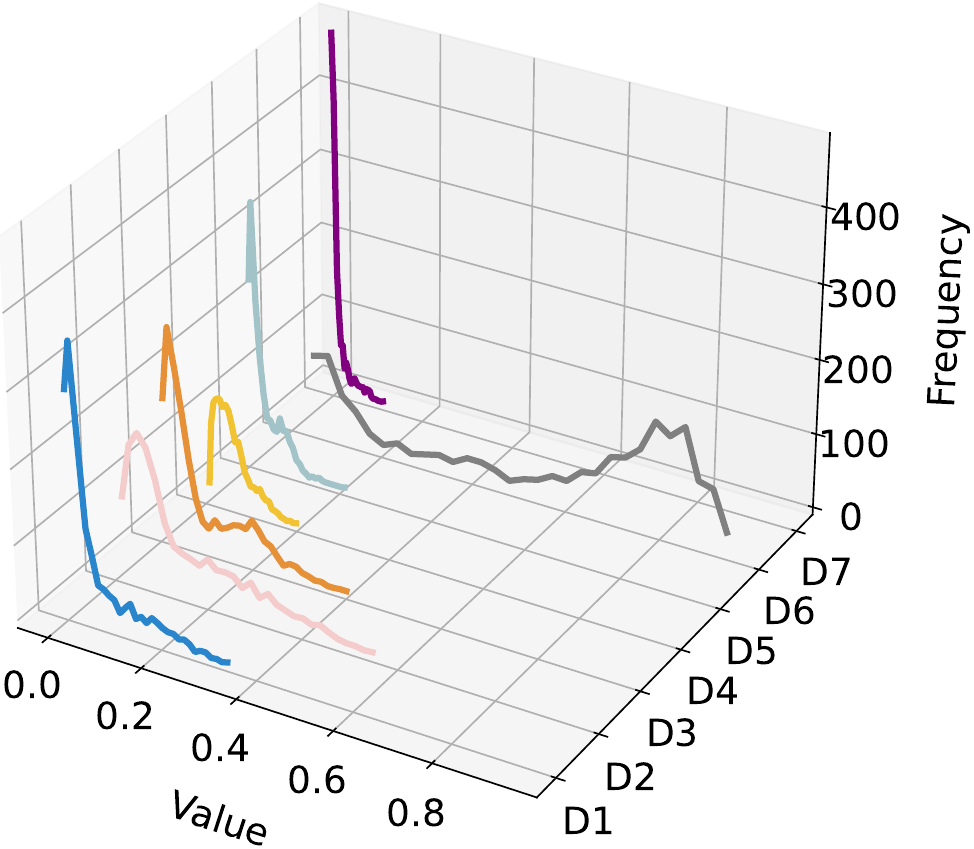} 
                \caption{STL10}
            \end{subfigure}
            \begin{subfigure}{0.2\textwidth}
                \centering
                \includegraphics[width=\linewidth]{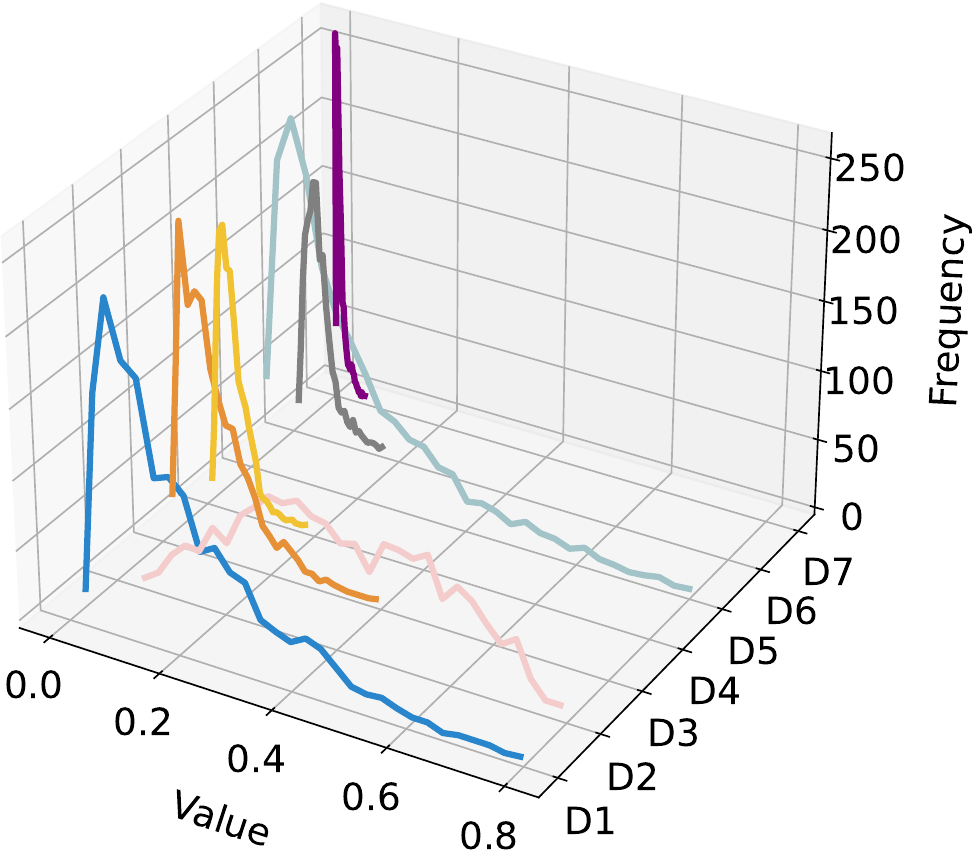} 
                \caption{SVHN}
            \end{subfigure}
            \begin{subfigure}{0.2\textwidth}
                \centering
                \includegraphics[width=\linewidth]{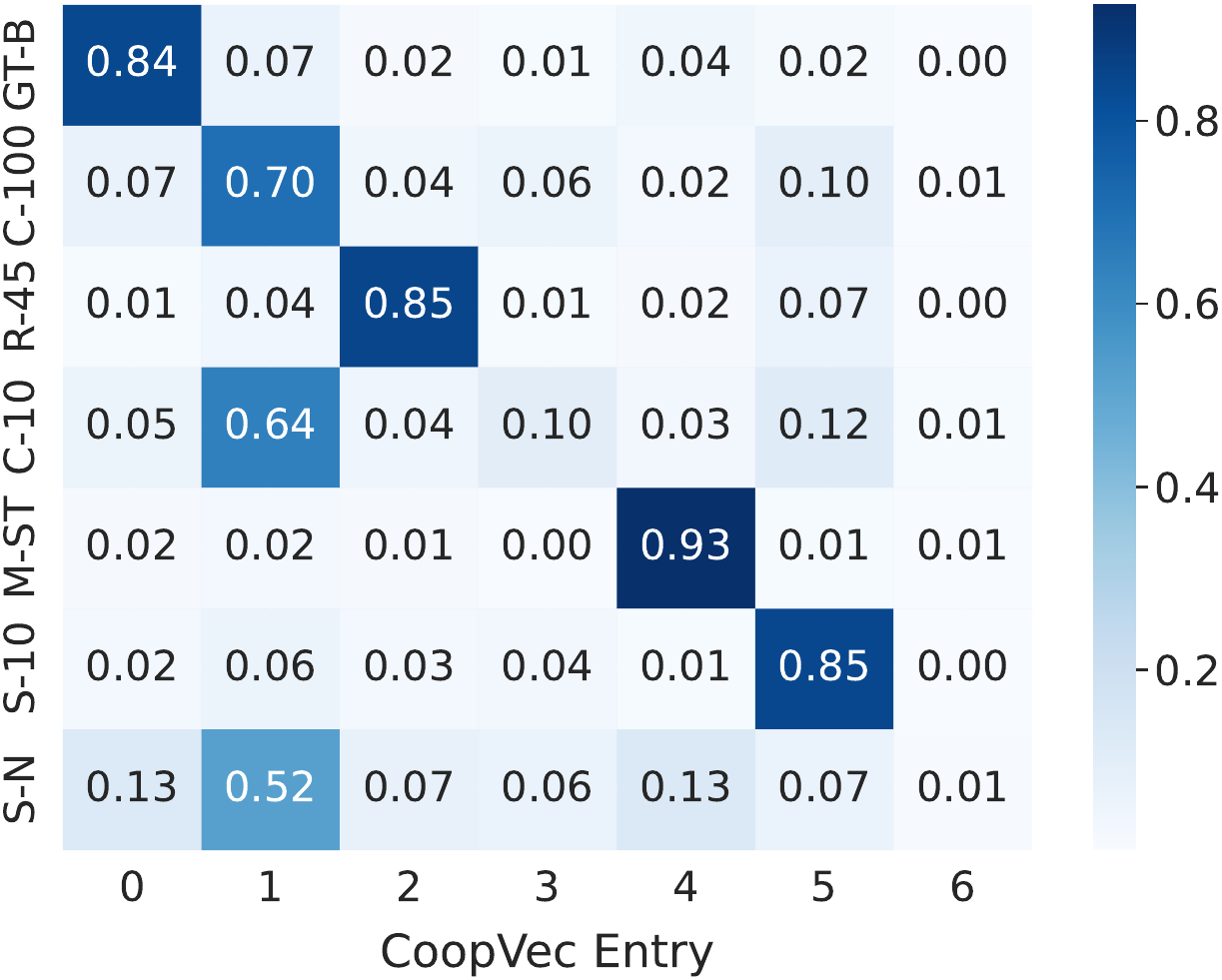} 
                \caption{CoopVec Map}
            \end{subfigure}
            
            \caption{(b) CLIP-ViT-B/32.}
        \end{subfigure}
    \end{minipage}
    \caption{CoopVec Distribution of seven different datasets and the corresponding CoopVec Map after training for one epoch. D1 to D7 represents the name of different datasets (D1-GTSRB, D2-CIFAR100, D3-RESISC45, D4-CIFAR10, D5-MNIST, D6-STL10, D7-SVHN).}
    \label{fig:coopvec_map_sup}
\end{figure*}

\begin{figure}[h]
    \centering
    \begin{minipage}{0.26\linewidth} 
        \centering
        \includegraphics[width=0.95\linewidth]{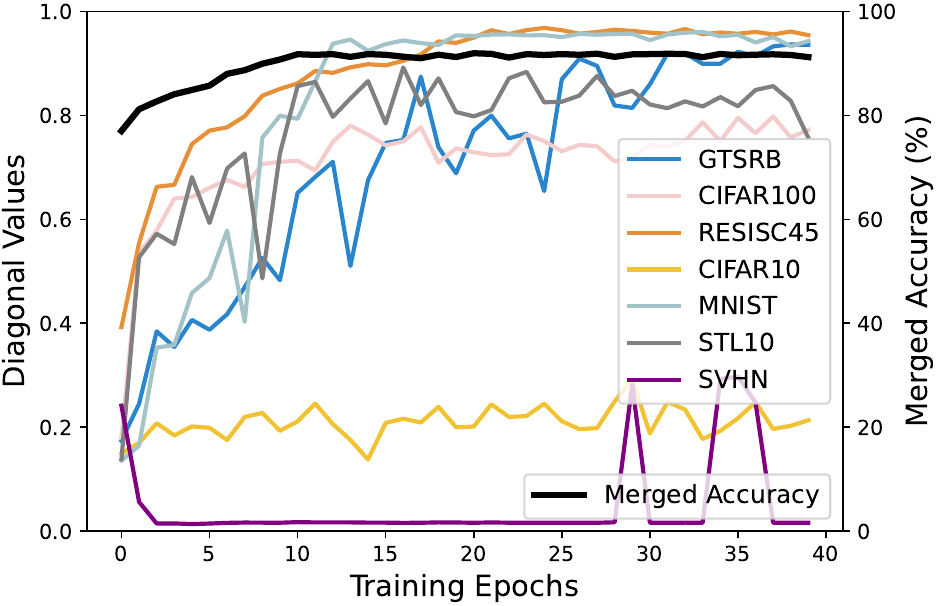}
        \includegraphics[width=0.95\linewidth]{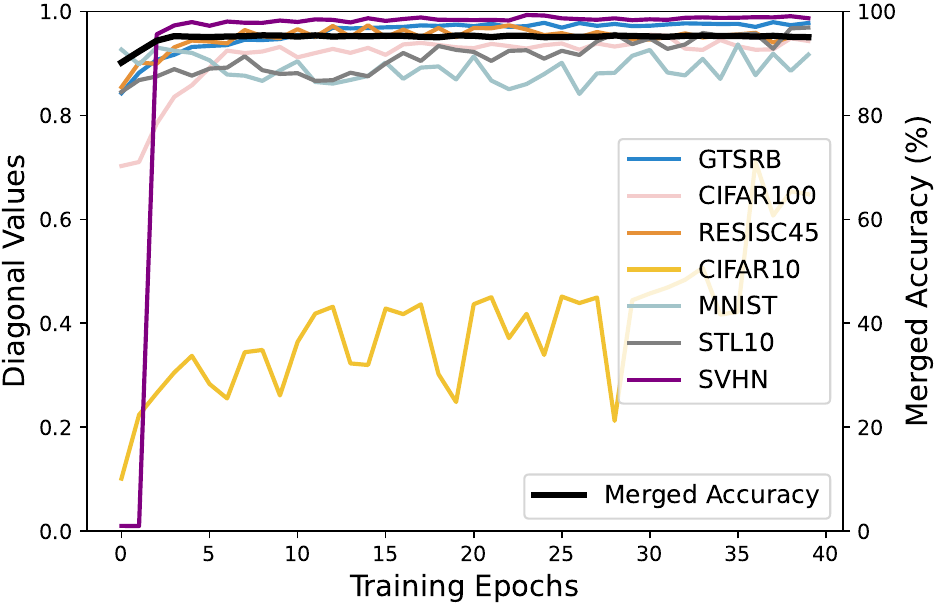}
        \caption{The variation of the diagonal values of CoopVec Map throughout the training process using CLIP-RN50 (top) and CLIP-ViT-B/32 (bottom).}
        \label{map_train_curve_sup}
    \end{minipage}%
    \hspace{0.03\linewidth} 
    \begin{minipage}{0.22\linewidth}
        \renewcommand{\arraystretch}{1.75} 
        \setlength{\tabcolsep}{4pt} 
        \centering
        \captionof{table}{Performance when using the CoopVec Map.}
        \begin{adjustbox}{width=\linewidth} 
            \begin{tabular}{c|c|c|c|c}
                \hline
                \multirow{2}{*}{\textbf{Dataset}} & \multicolumn{4}{c}{\textbf{Performance}} \\ \cline{2-5}
                & \textbf{Mer.}\textcolor{pink}{\ding{117}} & \textbf{Ens.}\textcolor{teal}{\ding{108}} & \textbf{Mer.}\textcolor{pink}{\ding{117}} & \textbf{Ens.}\textcolor{teal}{\ding{108}} \\ \hline
                \multirow{2}{*}{GTSRB} & \multicolumn{2}{c|}{CLIP-RN50} & \multicolumn{2}{c}{CLIP-ViT-B/32} \\ \cline{2-5}
                & 96.74 & 94.88 & 98.86 & 98.84 \\ \hline
                \multirow{2}{*}{CIFAR100} & \multicolumn{2}{c|}{CLIP-RN50} & \multicolumn{2}{c}{CLIP-ViT-B/32} \\ \cline{2-5}
                & 76.84 & 77.33 & 85.97 & 85.59 \\ \hline
                \multirow{2}{*}{RESISC45} & \multicolumn{2}{c|}{CLIP-RN50} & \multicolumn{2}{c}{CLIP-ViT-B/32} \\ \cline{2-5}
                & 92.06 & 92.54 & 93.63 & 93.84 \\ \hline
                \multirow{2}{*}{CIFAR10} & \multicolumn{2}{c|}{CLIP-RN50} & \multicolumn{2}{c}{CLIP-ViT-B/32} \\ \cline{2-5}
                & 91.98 & 92.51 & 95.84 & 96.37 \\ \hline
                \multirow{2}{*}{MNIST} & \multicolumn{2}{c|}{CLIP-RN50} & \multicolumn{2}{c}{CLIP-ViT-B/32} \\ \cline{2-5}
                & 99.50 & 99.64 & 99.57 & 99.57 \\ \hline
                \multirow{2}{*}{STL10} & \multicolumn{2}{c|}{CLIP-RN50} & \multicolumn{2}{c}{CLIP-ViT-B/32} \\ \cline{2-5}
                & 89.29 & 92.95 & 96.21 & 96.06 \\ \hline
                \multirow{2}{*}{SVHN} & \multicolumn{2}{c|}{CLIP-RN50} & \multicolumn{2}{c}{CLIP-ViT-B/32} \\ \cline{2-5}
                & 95.78 & 95.19 & 96.69 & 96.67 \\ \hline
                \multirow{1}{*}{Avg.Acc} & \makecell{91.74\\ \textcolor{teal}{(+34.83)}} & \makecell{92.15\\ \textcolor{teal}{(+3.57)}} & \makecell{95.25\\ \textcolor{teal}{(+14.92)}} & \makecell{95.28\\ \textcolor{teal}{(+3.45)}} \\ \hline
            \end{tabular}
            \label{map_table_perf_sup}
        \end{adjustbox}
    \end{minipage}
    \vspace{-5mm}
\end{figure}

\begin{table}[ht]
    \centering
    \resizebox{0.5\linewidth}{!}{
        \begin{tabular}{c|c|c|c|c|c|c}
            \hline
            \textbf{Training Epoch} & \textbf{Type} & \textbf{1} & \textbf{2} & \textbf{3} & \textbf{4} & \textbf{5} \\ 
            \hline
            \multirow{2}{*}{CLIP-ViT-B/32 (Ori.)} & 
             \textcolor{pink}{\ding{117}}Mer.& 94.32 & 93.38  & 94.84 & 94.81 & 94.67  \\ \cline{2-7} 
            & \textcolor{teal}{\ding{108}}Ens. & 94.94 & 94.90 & 95.13 & 95.05 & 95.05 \\
            \hline
            \multirow{2}{*}{CLIP-ViT-B/32 (Trans.)} & 
            \textcolor{pink}{\ding{117}}Mer.& 93.65 & 93.81 & 94.08 & 94.27 & 94.33 \\ \cline{2-7} 
            & \textcolor{teal}{\ding{108}}Ens. & 94.34 & 94.26 & 94.21 & 94.43 & 94.42 \\
            \hline
            \textbf{Training Epoch} & \textbf{Type} &  \textbf{6} & \textbf{7} & \textbf{8} & \textbf{9} & \textbf{10} \\ 
            \hline
            \multirow{2}{*}{CLIP-ViT-B/32 (Ori.)} & 
            \textcolor{pink}{\ding{117}}Mer.& 94.88 & 94.82 & 94.81 & 95.44 & 94.98  \\ \cline{2-7} 
            & \textcolor{teal}{\ding{108}}Ens. & 95.00 & 95.05 & 95.00 & 95.46 & 94.98  \\
            \hline
            \multirow{2}{*}{CLIP-ViT-B/32 (Trans.)} & 
            \textcolor{pink}{\ding{117}}Mer.& 94.16 & 94.36 & 94.02 & 94.11 & 94.01    \\ \cline{2-7} 
            & \textcolor{teal}{\ding{108}}Ens. & 94.25 & 94.35 & 94.30 & 94.20 & 94.26    \\
            \hline
        \end{tabular}
    }
    \caption{The transferability of \texttt{NeuLig}. Ori. refers to the original performance, while Trans. indicates the performance after directly applying Portland to the other group of models.}
    \label{tab:trans}
    \vspace{-3mm}
\end{table}

\section{CoopVec Map under More Models}
\label{app_co_map_more}
In Figure \ref{fig:coopvec_map_sup}, we depict the distribution of CoopVecs at the initial training stage for a seven-model collaboration, and the final CoopVec Map derived from this distribution, while Figure \ref{map_train_curve_sup} and Table \ref{map_table_perf_sup} capture the variation in the diagonal elements of the CoopVec Map throughout training, alongside the performance achieved using CoopVec for model collaboration. These experimental results clearly demonstrate that the conclusions presented in Section \ref{coop_map} of the main manuscript remain broadly valid and applicable in scenarios involving collaboration among a larger number of models.
   
\section{Transferability of Portland}
In this experiment, we split each dataset into two equal-sized subsets and train two separate models on each subset, referred to as model-A and model-B for simplicity. The objective is to explore whether the Portland trained collaboratively using all model-As, can be directly transferred to scenarios where all model-Bs collaborate, thereby assessing Portland's transferability. The results are summarized in Table \ref{tab:trans}. Notably, despite the fact that Portland was not explicitly trained for the model-Bs, the performance before and after the transfer remains largely consistent, highlighting Portland's robust transferability.

\section{Resilience of NeuLig With Varying Dataset Scales}
In the main manuscript, we explore the performance variation of \texttt{NeuLig} under different visible dataset scales when five models collaborate. We observe that performance consistency is well-maintained even with very small dataset scales. Here, consistent with Table \ref{tab:datalevel_main_7models}, we extend this investigation further by introducing two additional models, bringing the total model number to seven, which means we examine the impact of dataset scale on performance in the seven-model collaboration scenario. The results are presented in Table \ref{tab:data_scale_7models}. 

\begin{table}[h]
    \centering
    \resizebox{0.5\linewidth}{!}{
        \begin{tabular}{c|c|c|c|c|c|c}
            \hline
            \textbf{Data Scale} & \textbf{Type} & \textbf{0.01} & \textbf{0.05} & \textbf{0.1} & \textbf{0.15} & \textbf{0.2} \\ 
            \hline
            \multirow{2}{*}{CLIP-RN50} & \textcolor{pink}{\ding{117}}Mer.& 73.63 & 84.63 & 89.63 & 90.46 & 91.05 \\ \cline{2-7} 
            & \textcolor{teal}{\ding{108}}Ens. & 85.02 & 86.73 & 89.95 & 91.22 & 91.58 \\
            \hline
            \multirow{2}{*}{CLIP-ViT-B/32} & \textcolor{pink}{\ding{117}}Mer.& 88.74 & 93.68  & 94.91 & 94.91 & 95.23  \\ \cline{2-7} 
            & \textcolor{teal}{\ding{108}}Ens. & 93.17 & 94.71 & 95.10 & 95.19 & 95.24 \\
            \hline
            \textbf{Data Scale} & \textbf{Type} &  \textbf{0.3} & \textbf{0.4} & \textbf{0.6} & \textbf{0.8} & \textbf{1.0} \\ 
            \hline
            \multirow{2}{*}{CLIP-RN50} & \textcolor{pink}{\ding{117}}Mer.& 91.95 & 92.04 & 92.08 & 92.83 & 92.66    \\ \cline{2-7} 
            & \textcolor{teal}{\ding{108}}Ens. & 91.88 & 92.57 & 92.43 & 92.96 & 92.85 \\
            \hline
            \multirow{2}{*}{CLIP-ViT-B/32} & \textcolor{pink}{\ding{117}}Mer.& 95.30 & 95.00 & 94.92 & 94.91 & 95.43  \\ \cline{2-7} 
            & \textcolor{teal}{\ding{108}}Ens. & 95.18 & 95.23 & 95.26 & 95.26 & 95.43  \\
            \hline
        \end{tabular}
    }
    \caption{The performance variation of \texttt{NeuLig} under the semi-supervised learning setup when datasets of different scales are used.}
    \label{tab:data_scale_7models}
\end{table}

\begingroup
\xspaceskip=3pt    
Aligned with the main manuscript, we evaluate 10 different dataset scales to cover a wide range of conditions. The experimental results reveal that \texttt{NeuLig} demonstrates strong resilience to data scale, even in scenarios involving a larger number of collaborating models. Remarkably, even under extreme conditions (e.g., at a scale of 0.1), it continues to achieve performance consistency and maintain superior performance.
\par
\endgroup

{
    \small
    \bibliographystyle{ieeenat_fullname}
    \bibliography{main}
}

\end{document}